\documentclass[10pt,twocolumn,letterpaper]{article}

\usepackage[pagenumbers]{cvpr} 
\usepackage[accsupp]{axessibility}

\usepackage[dvipsnames]{xcolor}

\usepackage{algorithm}
\usepackage{algorithmic}
\usepackage{booktabs}       
\usepackage{amsfonts}       
\usepackage{url}            
\usepackage{nicefrac}       
\usepackage{microtype}      
\usepackage{multirow}
\usepackage{multicol}
\usepackage{xspace}
\usepackage{adjustbox}
\usepackage{import}
\usepackage{subcaption}
\usepackage{color, colortbl}
\usepackage{makecell}
\usepackage{enumitem}
\usepackage{wasysym}

\usepackage{bbding}
\usepackage{pifont}
\usepackage{listings}
\usepackage{tabulary}
\usepackage{diagbox}
\usepackage[dvipsnames]{xcolor}
\usepackage{dirtree}
\usepackage{tikz}
\usepackage{comment}
\usepackage{amsmath}
\usepackage{lineno}

\newcommand{\gr}[1]{\small{\textcolor{ForestGreen}{(\textbf{#1})}}}
\newcommand{\bgr}[1]{{\textcolor{ForestGreen}{(\textbf{#1})}}}
\newcommand{\cmark}{\ding{51}}%
\newcommand{\xmark}{\ding{55}}%
\newcommand{\sm}{\emph{supplementary material}}

\definecolor{baselinecolor}{gray}{.9}

\definecolor{demphcolor}{RGB}{144,144,144}

\definecolor{l_color}{RGB}{0,0,0}
\definecolor{f_color}{RGB}{0,0,0}
\newcommand{\demph}[1]{\textcolor{demphcolor}{#1}}

\definecolor{cvprblue}{rgb}{0.21,0.49,0.74}
\usepackage[pagebackref,breaklinks,colorlinks,citecolor=cvprblue]{hyperref}

\title{Enhancing Visual Continual Learning with Language-Guided Supervision}

\author{
Bolin Ni$^{1,2^\star}$, Hongbo Zhao$^{1,2^\star}$, Chenghao Zhang$^{1,2}$, Ke Hu$^{2}$\\
Gaofeng Meng$^{1,2,3\dagger}$, Zhaoxiang Zhang$^{1,2,3}$, Shiming Xiang$^{1,2}$\\
{\small $^{1}$State Key Laboratory of Multimodal Artificial Intelligence Systems, Institute of Automation, Chinese Academy of Sciences.} \\
{\small $^{2}$School of Artificial Intelligence, University of Chinese Academy of Sciences.} \\
{\small $^{3}$Centre for Artificial Intelligence and Robotics, HK Institute of Science \& Innovation, Chinese Academy of Sciences.} \\
\small{\texttt{nibolin2019@ia.ac.cn, gfmeng@nlpr.ia.ac.cn}}
}

\begin{document}

\twocolumn[{
\renewcommand\twocolumn[1][]{#1}
\maketitle
\begin{center}
    \centering
    \includegraphics[width=2.1\columnwidth]{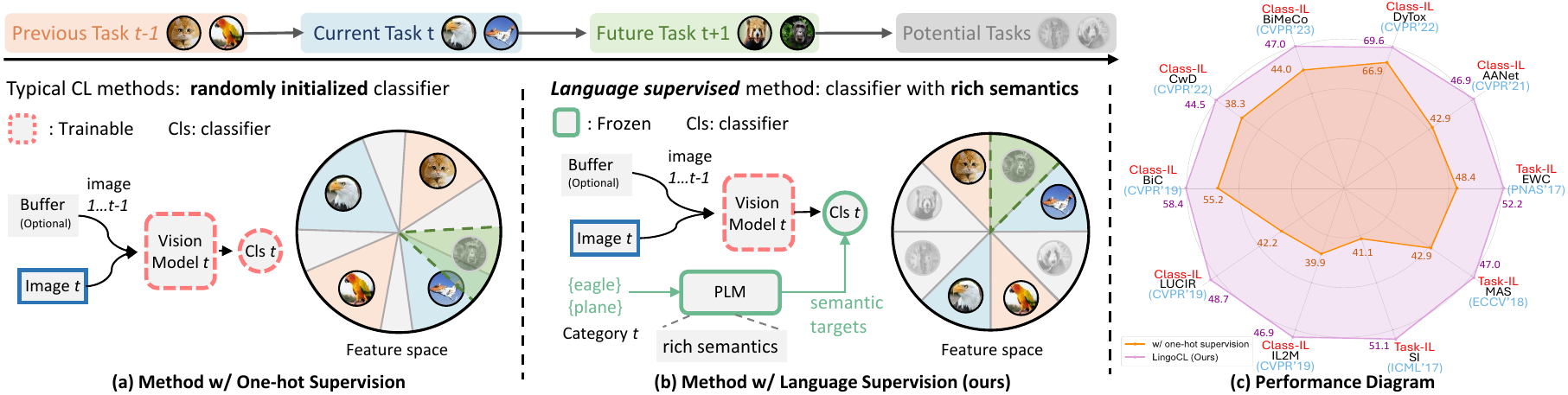}
    \captionof{figure}{We introduce LingoCL, a simple yet effective continual learning paradigm leveraging language-guided supervision, which can be integrated into most existing approaches seamlessly. (a) Overview of the typical methods which are supervised only by one-hot labels. (b) Overview of the proposed LingoCL which is supervised by semantic targets generated from the pretrained language model. (c) LingoCL is versatile, which significantly enhances the performance of mainstream methods in class-, task- and domain-incremental scenarios.} 
    \label{fig:overview}
\end{center}
}]

\maketitle
\renewcommand{\thefootnote}{}
\footnotetext{$^\star$Equal contribution. $^\dagger$Corresponding author.
}

\begin{abstract}
Continual learning (CL) aims to empower models to learn new tasks without forgetting previously acquired knowledge. Most prior works concentrate on the techniques of architectures, replay data, regularization, \etc. However, the category name of each class is largely neglected. Existing methods commonly utilize the one-hot labels and randomly initialize the classifier head. We argue that the scarce semantic information conveyed by the one-hot labels hampers the effective knowledge transfer across tasks. In this paper, we revisit the role of the classifier head within the CL paradigm and replace the classifier with semantic knowledge from pretrained language models (PLMs). Specifically, we use PLMs to generate semantic targets for each class, which are frozen and serve as supervision signals during training. Such targets fully consider the semantic correlation between all classes across tasks. Empirical studies show that our approach mitigates forgetting by alleviating representation drifting and facilitating knowledge transfer across tasks. The proposed method is simple to implement and can seamlessly be plugged into existing methods with negligible adjustments. Extensive experiments based on eleven mainstream baselines demonstrate the effectiveness and generalizability of our approach to various protocols. For example, under the class-incremental learning setting on ImageNet-100, our method significantly improves the Top-1 accuracy by 3.2\% to 6.1\% while reducing the forgetting rate by 2.6\% to 13.1\%.
\end{abstract}    
\vspace{-0.5cm}
\section{Introduction}
\label{sec:intro}

The main challenge in continual learning (CL) is \textit{catastrophic forgetting}, where models experience significant performance degradation on earlier tasks when new tasks are introduced. To address this, researchers have developed various strategies, including architecture-based~\cite{learntogrow,pnn,der,aanet}, replay-based~\cite{aljundi2019gradient,gdumb,cwd}, distillation-based~\cite{Rebuffi2017iCaRLIC,Douillard2020PODNetPO,lucir}, and regularization-based methods~\cite{ewc,rwalk,si}, alongside other notable contributions~\cite{l2p,khan2023introducing}.

However, most existing approaches overlook the significance of the semantic knowledge contained in category names. The prevailing trend in prior work leans towards using one-hot labels, coupled with the randomly initialized classifier head, and optimizing the encoder and classifier head jointly. Such a methodology is de facto paradigm for stationary environments. Nevertheless, in the CL scenarios, this practice presents two issues. Firstly, the problem of \textit{representation drifting} emerges. When the model encounters new tasks, the feature space could drift or even be overwritten, compromising the \textit{stability} of models. This drift arises because the optimization of the semantic target\footnote{Each row of the classifier's weights represents the semantic target for its corresponding class.} of each class is narrowly focused on its current task. Due to the limited access to old data and the unpredictability of new data, the model struggles to be compatible with the previous and future classes. For example, as shown in Fig.~\ref{fig:overview}(a), the potential future class ``chimpanzee" may erase the feature space of the learned class ``plane", exacerbating the forgetting of old tasks. Secondly, this particularity of data in CL also results in \textit{inefficient knowledge transfer}. Since the semantic targets in the classifier are randomly initialized without any prior knowledge, and are then optimized within individual tasks, it struggles to capture the semantic correlation across all tasks. This incompleteness in semantic correlations impedes the model's knowledge transfer, thereby affecting its \textit{plasticity}.

In this work, we study how to enhance CL performance by leveraging the semantic knowledge in category names from a classifier perspective. Inspired by the impressive generalization capabilities of pretrained language models (PLMs)~\cite{clip,gpt3}, we propose a simple yet effective approach, language-guided supervision for CL (LingoCL), which employs PLM to generate the semantic targets. Specifically, for the incoming task, we first use the category name of each class as input to the language model and take the outputs as the weights in the classifier. Then, the classifier is kept frozen during CL training, guiding the learning of the encoder. 
Our approach is motivated by the rich knowledge and strong generalization abilities of PLMs. Even with the limitations in previous and future data, PLMs ensure that each generated semantic target implicitly considers the semantic correlations between all classes. Therefore, these targets can be used to direct the learning of the encoder.
For instance, as illustrated in Fig.~\ref{fig:overview}(b), PLMs can provide the prior knowledge that the ``eagle'' in current tasks shares a similar semantic target with the learned ``parrot'' class, facilitating the knowledge transfer between the learned classes and new classes. We explore two types of language models in this work: self-supervised models on unimodal data and vision-supervised models on multimodal data. Our results demonstrate that both types of models can serve as excellent classifier heads, constantly improving performance. Moreover, the analysis in Sec.~\ref{sec:analysis} demonstrates that the improvements come from alleviating the representation drift and facilitating knowledge transfer, instead of the individual gains at each task.

Without loss of generality, we choose eleven methods as baselines and incorporate the text-supervised classifiers for them. Comprehensive experiments demonstrate the proposed methods are generally effective. In particular, under the class-incremental learning setting, LingoCL can improve the accuracy on ImageNet-100 by $3.2\%$ to $6.1\%$, and reduce the forgetting rate by $2.6\%$ to $13.1\%$. In task- and domain-incremental learning, LingoCL improves the accuracy by $3.9\%$ to $9.7\%$ and $1.2\%$ to $4.0\%$, respectively.

The contributions can be summarized as follows:
\begin{itemize}
    \item We point out that the semantic knowledge in category names is largely neglected by existing methods when initializing classifiers, which could have two issues, \ie, representation drifting and insufficient knowledge transfer.
    \item We propose LingoCL, a new CL paradigm with language-guided supervision. With the rich semantic knowledge in PLMs, we alleviate the abovementioned issues and thus enhance the performance of mainstream CL methods.
    \item The proposed LingoCL has several key advantages: 1) computation efficiency; 2) orthogonality to existing methods; 3) flexibility with various PLMs; and 4) versatility in diverse CL scenarios. Extensive experiments are conducted to systematically examine our method.
\end{itemize}

\section{Related Work}
\label{sec:related_work}

\textbf{Continual learning.} To alleviate catastrophic forgetting, researchers have explored various routes. 
\textit{Regulation-based methods}~\cite{ewc,si,mas,rwalk} aim to prevent catastrophic forgetting by penalizing the changes of network parameters when learning current tasks. \textit{Replay-based methods} entail selecting a subset of data from previous tasks~\cite{aljundi2019gradient,gdumb,zhao2024continual} or using generative models to produce synthetic data~\cite{hu2019overcoming,fearnet,ostapenko2019learning} as ``replayed" data to preserve the knowledge of previous tasks. \textit{Distillation-based methods} take the model trained on the previous task as the teacher to supervise the learning of the current model. These methods can be divided into logits distillation~\cite{Rebuffi2017iCaRLIC,bic}, feature distillation~\cite{Douillard2020PODNetPO,lucir}, and relational distillation~\cite{tao2020topology,tao2020few}. \textit{Architecture-based methods} involve dynamic allocation of different parameters for each task through architecture expansion~\cite{learntogrow,pnn,dytox,der} or mask operation~\cite{packnet,piggyback}. 
\textit{Rectification-based methods} analyzes the abnormal behaviors in CL models compared to oracle models and tries to rectify them. These methods usually focus on the imbalance in the feature embedding~\cite{il2m,aanet,cwd} or network weights~\cite{bic,e2e,wa}.

Most existing methods commonly use one-hot labels coupled with randomly initialized classifiers, ignoring the category names seriously. In contrast to them, our work studies whether and how to improve CL by leveraging the semantic information contained in the category names.

\noindent\textbf{Cross-modality adaptation.}
In recent years, transferring language knowledge to visual modeling has emerged as a new paradigm. For example, contrastive language-image pretraining demonstrates impressive ``zero-shot” transfer and generalization capacities~\cite{clip,align,florence}. Moreover, some works explore how to model the vision input using pretrained language models in order to transfer the ability of language models~\cite{flamingo,frozen,blip2}. Another line of work focuses on how to improve the vision encoder with the guidance of the language information~\cite{ni2022expanding}. For instance, Tex~\cite{wang2023improved} proposes to use language models to reduce the bias in the classifier of fine-tuned visual models. DUET~\cite{chen2023duet} integrates the latent semantic knowledge from PLMs to vision models for better zero-shot recognition ability. Additionally, Lei \textit{et al}.~\cite{vision_bring_to} designed a suite of evaluation tasks across various perception aspects and showed that language models can learn visual features from vast amounts of data, including shape, texture, and color, and that vision supervision can enhance the comprehension of visual concepts. In this work, we are pioneering the exploration of how to transfer knowledge in language models to address the catastrophic forgetting issue in continual learning.

\section{Methodology}
\label{sec:approach}
\begin{figure}
\centering
    \begin{subfigure}{0.23\textwidth}
        \includegraphics[width=1\textwidth]{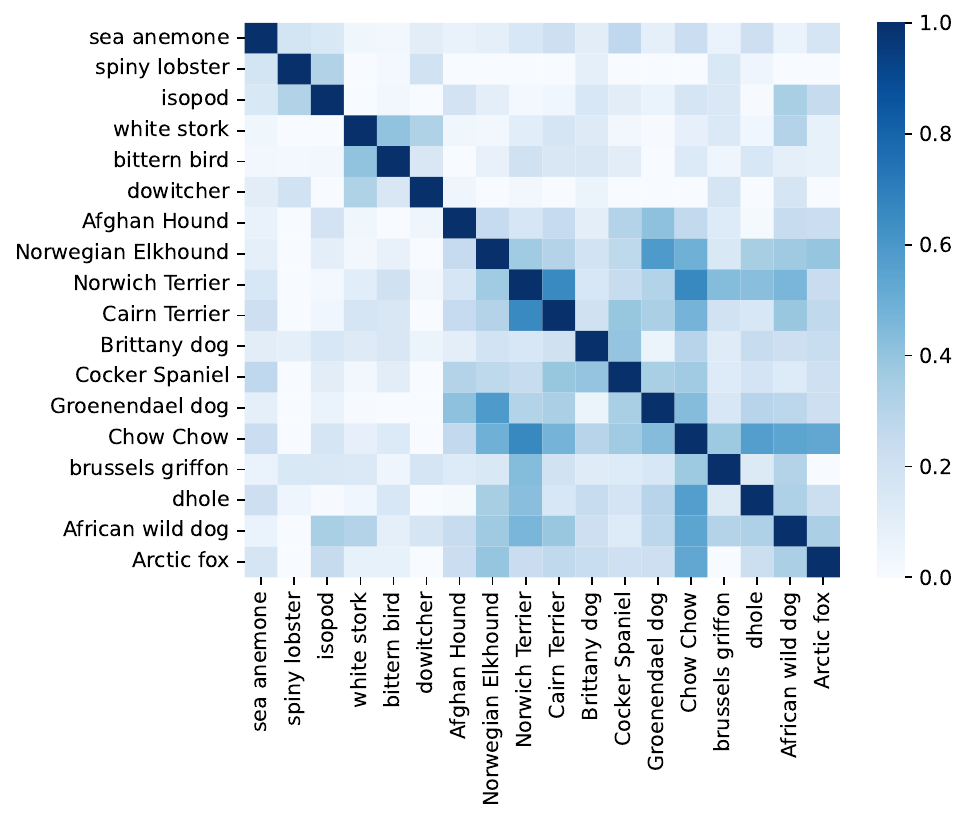}
        \vspace{-0.7cm}
	\caption{w/ one-hot supervision}
    \end{subfigure}
    \begin{subfigure}{0.23\textwidth}
        \includegraphics[width=1\textwidth]{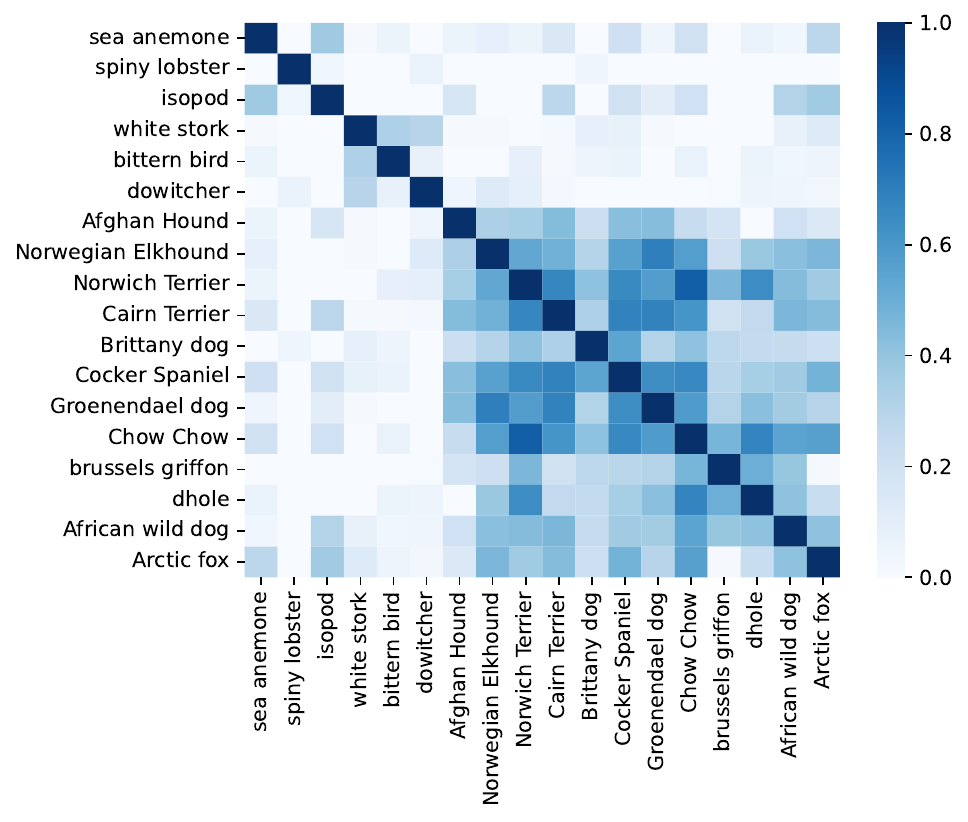}
        \vspace{-0.7cm}
	\caption{w/ language supervision (ours)}
    \end{subfigure}
    \caption{Comparison of the inter-class correlation maps. LingoCL facilitates more efficient knowledge transfer among similar classes.} \label{fig:corr}
\end{figure}
\begin{figure}
\hspace*{-0.5cm} 
\centering
    \includegraphics[width=0.45\textwidth]{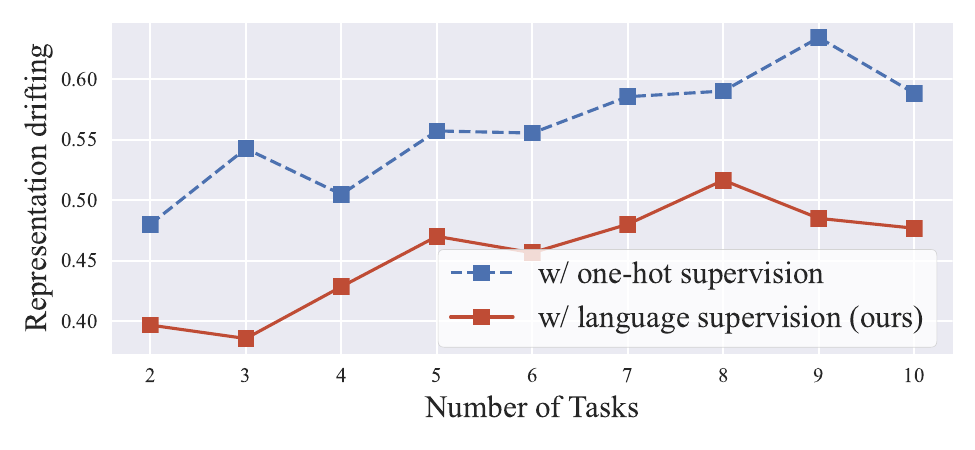}
    \caption{Quantitative analysis of representation drifting on ImageNet-100 with 10 tasks. LingoCL effectively alleviates the representation drifting in the CL process.}
    \vspace{-0.3cm}
    \label{fig:drifting}
\end{figure}

\subsection{Revisiting Classifier in Existing CL Paradigms}
We first review the typical CL paradigms from a classifier perspective. In CL scenarios, the vision encoder, denoted by $g_V$, is sequentially optimized over tasks. Each task typically requires an individual classifier head. For $t$-th task, we symbolize the training dataset as $\mathcal{D}_t = \{(\boldsymbol{x}_t, y_t)\}$ that contains $C_t$ disjoint classes and the classifier head as $\mathbf{W}_t \in \mathbb{R}^{C_t\times d}$. 
The vision encoder $g_V$ and semantic targets in $\mathbf{W}_t$ are optimized jointly, following the learning objective in the stationary environment:
\begin{equation}
    \textcolor{l_color}{g^*_V}, \textcolor{l_color}{\mathbf{W}_t^*} = \underset{\textcolor{l_color}{\Theta_V},\textcolor{l_color}{\mathbf{W}_t}}{\mathrm{arg min}}~
                {\mathbb{E}}_{{\boldsymbol{x}_t,y_t \sim \mathcal{D}_t}}\big[ \mathcal{L}(\mathrm{sim}(\textcolor{l_color}{\mathbf{W}_t}, \textcolor{l_color}{g_V}(\boldsymbol{x}_t)), y_t) \big], \label{eq:cl_paradigm}
\end{equation}
where $\mathbf{W}_t \sim \mathcal{N}(\mathbf{0}, \mathbf{I}_d)$. The function $\mathrm{sim}(\mathbf{W}, g_V(\boldsymbol{x}_t))$ calculates the similarity between the image embedding $g_\theta(\boldsymbol{x})$, and every semantic target in \textcolor{black}{$\mathbf{W}_t$}, using the inner product.

However, this approach encounters challenges specific to CL. Firstly, CL benefits from knowledge transfer across tasks, such as forward and backward transfer, but randomly initialized classifiers struggle to capture the semantic similarity among classes across tasks, resulting in \textit{inefficient knowledge transfer}. Secondly, the optimization of each semantic target is confined to its current task. This narrow focus overlooks the broader compatibility between semantic targets spanning all tasks. Such a shortsighted learning approach can cause conflicts between the semantic targets of different tasks, inducing \textit{representation drifting or erasure} in the feature space.

\subsection{Our Proposed Language-Guided Supervision} \label{sec:our_method}
To address these issues, we utilize the rich semantic knowledge contained in pretrained language models to guide the learning process for each task. Specifically, for an incoming task $t$, the procedure is as follows:
\begin{enumerate}[label=(\roman*)]
  \item  Gathering the category names $[\boldsymbol{l}_1,\cdots,\boldsymbol{l}_{C_t}]$ of task $t$. 
  \item  Feeding these category names into PLM to generate the semantic targets for the classifier \textcolor{f_color}{$\tilde{\mathbf{W}}_t$}:
    \begin{equation}
    \textcolor{f_color}{\tilde{\mathbf{W}}_t} = \textcolor{f_color}{g_T}([\boldsymbol{l}_1,\cdots,\boldsymbol{l}_{C_t}]),
    \end{equation}
  \item  Optimizing the vision encoder \textcolor{l_color}{$g_V$}, while keeping the classifier \textcolor{f_color}{$\tilde{\mathbf{W}}_t$} \textit{frozen}:
    \begin{equation}
        \textcolor{l_color}{g^*_V} = \underset{\textcolor{l_color}{\Theta_V}}{\mathrm{arg min}}~
                    {\mathbb{E}}_{{\boldsymbol{x}_t,y_t \sim \mathcal{D}_t}}\big[ \mathcal{L}(\mathrm{sim}(\textcolor{f_color}{\tilde{\mathbf{W}}_t}, \textcolor{l_color}{g_V}(\boldsymbol{x}_t)), y_t) \big].
    \end{equation}
\end{enumerate}
The classifier \textcolor{f_color}{$\tilde{\mathbf{W}}_t$} is kept frozen to preserve the semantic knowledge from being disturbed or forgotten in the CL process. As these generated semantic targets are optimized with sufficient data and concepts, they effectively serve as supervision signals, directing the vision encoder's optimization.

In light of the outlined methodology, LingoCL has several key advantages: 1) it is \textit{computation-efficient}; leveraging category names requires only a single forward propagation, with a negligible cost comparison to overall training; 2) it provides flexibility to utilize knowledge from \textit{various} language models, promoting easy integration of the latest PLM advancements; 3) it is \textit{orthogonal} to most of existing CL methods, allowing for seamless integration; 4) it is \textit{versatile}, and compatible with diverse CL scenarios such as class-, task- and domain-IL.

\setlength{\tabcolsep}{1.0mm}{
\begin{table*}[t!]
  \centering
  \resizebox{\textwidth}{!}{
  \begin{tabular}
  {p{2.5cm}p{2.5cm}p{1.5cm}p{1.5cm}p{1.3cm}p{0.cm}p{1.5cm}p{1.5cm}p{1.3cm}p{0.cm}p{1.5cm}p{1.5cm}p{1.3cm}}
  \toprule
   \multirow{3}{*}{Method} & \multirow{3}{*}{Backbone} & \multicolumn{3}{c}{\emph{$B$=10, $C$=10}} && \multicolumn{3}{c}{\emph{$B$=5, $C$=5}} && \multicolumn{3}{c}{\emph{$B$=2, $C$=2}} \\
  \cmidrule{3-5} \cmidrule{7-9} \cmidrule{11-13} 
   & & Avg ($\uparrow$) & Last ($\uparrow$) & $\mathcal{F}$ ($\downarrow$) && Avg ($\uparrow$) & Last ($\uparrow$) & $\mathcal{F}$ ($\downarrow$) &&  Avg ($\uparrow$) & Last ($\uparrow$) & $\mathcal{F}$ ($\downarrow$) \\
    \midrule
    Oracle  & ResNet-18 & 77.6 & 77.6 & - && 77.6 & 77.6 & - && 77.6 & 77.6 & -  \\
    \quad w/ LingoCL & & 78.0 & 78.0 & - && 78.0 & 78.0 & - && 78.0 & 78.0 & - \\
    \midrule
    \multicolumn{5}{l}{\textit{Architecture-based methods}} \\
    AANet~\cite{aanet} & ResNet-18  & 64.6\scriptsize{$\pm$0.2} & 49.1\scriptsize{$\pm$0.2} & - && 62.5\scriptsize{$\pm$0.3} & 42.5\scriptsize{$\pm$0.3} & - && 57.7\scriptsize{$\pm$0.5} & 37.6\scriptsize{$\pm$0.5} & - \\

    \quad w/ LingoCL && \textbf{65.4}\scriptsize{$\pm$0.1} \gr{+0.8} & \textbf{50.4}\scriptsize{$\pm$0.1} \gr{+1.3} & - && \textbf{63.2}\scriptsize{$\pm$0.2} \gr{+0.7} & \textbf{44.5}\scriptsize{$\pm$0.4} \gr{+2.0} & - && \textbf{58.7}\scriptsize{$\pm$0.4} \gr{+1.0} & \textbf{38.6}\scriptsize{$\pm$0.5} \gr{+1.0} & -  \\

    DyTox~\cite{dytox} & ConViT & 69.5\scriptsize{$\pm$0.0} & 52.8\scriptsize{$\pm$0.2} & 33.0\scriptsize{$\pm$0.0} && 67.4\scriptsize{$\pm$0.1} & 48.1\scriptsize{$\pm$0.3} & 37.8\scriptsize{$\pm$0.0} && 64.5\scriptsize{$\pm$0.2} & 44.8\scriptsize{$\pm$0.3} & 41.3\scriptsize{$\pm$0.1} \\
    \quad w/ LingoCL && \textbf{71.9}\scriptsize{$\pm$0.0} \gr{+2.4} & \textbf{58.9}\scriptsize{$\pm$0.1} \gr{+6.1} & \textbf{24.9}\scriptsize{$\pm$0.0} \gr{-8.1} && \textbf{70.0}\scriptsize{$\pm$0.1} \gr{+2.6} & \textbf{52.3}\scriptsize{$\pm$0.3} \gr{+4.2} & \textbf{30.5}\scriptsize{$\pm$0.1} \gr{-7.3} && \textbf{65.9}\scriptsize{$\pm$0.1} \gr{+1.4} & \textbf{46.3}\scriptsize{$\pm$0.3} \gr{+1.5} & \textbf{36.1}\scriptsize{$\pm$0.2} \gr{-5.2} \\

    BiMeCo~\cite{bimeco} & MobileNet-V2$^\dagger$ & 59.7\scriptsize{$\pm$0.5} & 45.9\scriptsize{$\pm$0.5} & 35.4\scriptsize{$\pm$0.6} && 52.5\scriptsize{$\pm$0.4} & 38.7\scriptsize{$\pm$0.3} & 43.5\scriptsize{$\pm$0.7} && 37.2\scriptsize{$\pm$1.0} & 25.9\scriptsize{$\pm$1.3} & 50.7\scriptsize{$\pm$1.1} \\
    \quad w/ LingoCL && \textbf{60.2}\scriptsize{$\pm$0.2} \gr{+0.5} & \textbf{46.4}\scriptsize{$\pm$0.1} \gr{+0.5} & \textbf{32.3}\scriptsize{$\pm$0.1} \gr{-3.1} && \textbf{54.2}\scriptsize{$\pm$0.7} \gr{+1.7} & \textbf{39.1}\scriptsize{$\pm$0.6} \gr{+0.4} & \textbf{41.8}\scriptsize{$\pm$0.5} \gr{-1.7} && \textbf{43.4}\scriptsize{$\pm$0.9} \gr{+6.2} & \textbf{32.5}\scriptsize{$\pm$0.8} \gr{+6.6} & \textbf{45.0}\scriptsize{$\pm$0.9} \gr{-5.7} \\
    
    \midrule
    \multicolumn{5}{l}{\textit{Distillation-based methods}} \\
    LUCIR~\cite{lucir} & ResNet-18  & 60.2\scriptsize{$\pm$0.4} & 46.5\scriptsize{$\pm$0.7} & 37.3\scriptsize{$\pm$0.5} && 54.8\scriptsize{$\pm$0.7} & 41.7\scriptsize{$\pm$1.0} & 42.0\scriptsize{$\pm$0.5} && 45.6\scriptsize{$\pm$1.1} & 36.2\scriptsize{$\pm$1.6} & 44.5\scriptsize{$\pm$0.9} \\
    \quad  w/ LingoCL && \textbf{61.9}\scriptsize{$\pm$0.3} \gr{+1.7} & \textbf{47.5}\scriptsize{$\pm$0.6} \gr{+1.0} & \textbf{36.5}\scriptsize{$\pm$0.1} \gr{-0.8} && \textbf{56.3}\scriptsize{$\pm$0.5} \gr{+1.5} & \textbf{44.3}\scriptsize{$\pm$0.4} \gr{+2.6} & \textbf{39.8}\scriptsize{$\pm$0.8} \gr{-2.2} && \textbf{46.8}\scriptsize{$\pm$1.2} \gr{+1.2} & \textbf{37.0}\scriptsize{$\pm$0.6} \gr{+0.8} & \textbf{42.3}\scriptsize{$\pm$1.0} \gr{-2.2} \\

    BiC~\cite{bic} & ResNet-18 & 57.8\scriptsize{$\pm0.9$} & 41.2\scriptsize{$\pm1.0$} & 26.7\scriptsize{$\pm1.0$} && 50.1\scriptsize{$\pm0.6$} & 34.7\scriptsize{$\pm0.1$} & 28.7\scriptsize{$\pm0.7$} && 38.1\scriptsize{$\pm1.0$} & 23.6\scriptsize{$\pm0.4$} & 38.6\scriptsize{$\pm0.9$} \\ 
    \quad w/ LingoCL && \textbf{60.1}\scriptsize{$\pm0.5$} \gr{+2.3} & \textbf{43.4}\scriptsize{$\pm1.0$} \gr{+2.2} & \textbf{20.5}\scriptsize{$\pm1.0$} \gr{-6.2} && \textbf{51.6}\scriptsize{$\pm0.6$} \gr{+1.5} & \textbf{36.6}\scriptsize{$\pm1.3$} \gr{+1.9} & \textbf{19.6}\scriptsize{$\pm0.6$} \gr{-9.1} && \textbf{44.9}\scriptsize{$\pm1.0$} \gr{+6.8} & \textbf{31.1}\scriptsize{$\pm0.8$} \gr{+7.5} & \textbf{29.7}\scriptsize{$\pm0.6$} \gr{-8.9}\\

    \midrule
    \multicolumn{5}{l}{\textit{Rectification-based methods}} \\
    
    CwD~\cite{cwd} & ResNet-18 & 60.0\scriptsize{$\pm0.5$} & 46.7\scriptsize{$\pm0.5$} & 36.9\scriptsize{$\pm0.9$} && 54.4\scriptsize{$\pm0.6$} & 42.2\scriptsize{$\pm0.5$} & 41.5\scriptsize{$\pm0.4$} && 40.2\scriptsize{$\pm1.2$} & 34.0\scriptsize{$\pm1.2$} & 44.6\scriptsize{$\pm0.2$} \\
    \quad w/ LingoCL && \textbf{60.6}\scriptsize{$\pm0.7$} \gr{+0.6} & \textbf{47.6}\scriptsize{$\pm1.1$} \gr{+0.9} & \textbf{35.3}\scriptsize{$\pm1.0$} \gr{-1.3} && \textbf{55.7}\scriptsize{$\pm0.9$} \gr{+1.3} & \textbf{44.3}\scriptsize{$\pm0.5$} \gr{+2.1} & \textbf{39.2}\scriptsize{$\pm0.7$} \gr{-2.3} && \textbf{46.0}\scriptsize{$\pm0.7$} \gr{+5.8} & \textbf{38.4}\scriptsize{$\pm0.8$} \gr{+4.4} & \textbf{36.9}\scriptsize{$\pm0.8$} \gr{-7.7} \\
    
    IL2M \cite{il2m} & ResNet-18  & 57.8\scriptsize{$\pm0.3$} & 44.3\scriptsize{$\pm0.8$} & 41.0\scriptsize{$\pm0.3$} && 52.6\scriptsize{$\pm0.8$} & 40.5\scriptsize{$\pm0.4$} & 45.3\scriptsize{$\pm1.0$} && 44.0\scriptsize{$\pm1.1$} & 34.2\scriptsize{$\pm0.8$} & 48.5\scriptsize{$\pm0.5$} \\
    \quad w/ LingoCL && \textbf{62.1}\scriptsize{$\pm0.0$} \gr{+4.3} & \textbf{48.1}\scriptsize{$\pm1.0$} \gr{+3.8} & \textbf{37.8}\scriptsize{$\pm0.3$} \gr{-3.2} && \textbf{56.6}\scriptsize{$\pm0.4$} \gr{+4.0} & \textbf{44.0}\scriptsize{$\pm1.0$} \gr{+3.5} & \textbf{43.0}\scriptsize{$\pm1.2$} \gr{-2.3} && \textbf{48.0}\scriptsize{$\pm0.7$} \gr{+4.0} & \textbf{39.6}\scriptsize{$\pm0.0$} \gr{+5.4} & \textbf{42.8}\scriptsize{$\pm0.2$} \gr{-5.7} \\
    
  \bottomrule
\end{tabular}
}
\caption{Results on class-incremental experiments on CIFAR100 of Average accuracy (\%), last phase accuracy (\%) and forgetting rate $\mathcal{F}$ (\%) with and without text-supervised classifier at various CL settings. $B$ denotes the number of classes at the initial task, and $C$ denotes the number of classes in each task after the initial one. $\dagger$ denotes a modified version of the backbone as adapted by the original authors. Notably, for each metric, $\uparrow$ ($\downarrow$) indicates that the larger (the smaller) values, the better results are.}
\label{tab:cifar}
 \vspace{-0.2cm}
\end{table*}
}
\subsection{Quantitative Analysis} \label{sec:analysis}

Next, we examine our method to answer the two questions mentioned above: \textit{1) Does our method alleviate representation drifting,} and \textit{2) Does it facilitate knowledge transfer?}

To answer the first question, we perform a subspace analysis~\cite{anatomy} on challenging class-incremental learning protocol. Given the same input, let $\mathbf{F}_t, \mathbf{F}_{t'}\in \mathbb{R}^{n\times d}$ denote the output of the encoder after the $t$-th task and after the $t'$-th task ($t'>t$), respectively. $\mathbf{V}_{k,t}$ and $\mathbf{V}_{k,t'}$ are the top-$k$ principal directions of $\mathbf{F}_t$ and $\mathbf{F}_{t'}$, respectively. The representation drifting from the $t$-th task to the $t'$-th can be defined as:
\begin{equation}
    \mathrm{RepreDrift}_k(\mathbf{F}_t, \mathbf{F}_{t'}) = 1 - \frac{1}{k}\|\mathbf{V}_{k,t}^T \mathbf{V}_{k,t'} \|_{F}^2,
\end{equation}
where a smaller value indicates less representation drifting. We adopt LUCIR~\cite{lucir} as the baseline. Fig.~\ref{fig:drifting} shows the evolution of the first task's representations as the training progresses. LingoCL significantly reduces the representation drifting, demonstrating the capacity to enhance the stability of the CL model.

For the second question, we sample the first 18 classes in ImageNet-100 and calculate the inter-class correlation of the embeddings produced by the encoder with vanilla classifier and LingoCL. Note that these classes are scattered among different tasks. Fig.~\ref{fig:corr} shows that LingoCL exhibits certain inter-class correlations, indicating that a well-considered semantic target can facilitate knowledge transfer and improve the performance of CL. The above analyses offer an initial demonstration of the effectiveness of LingoCL. A more in-depth exploration of these issues is presented in Tab.~\ref{tab:component}.

\section{Experiments}\label{sec:experiments}
\begin{table*}[t]
  \centering
  \small
  \resizebox{1\textwidth}{!}{
  \begin{tabular}{lccccccccccccccc}
  \toprule
   \multirow{3}{*}{Method} & \multicolumn{3}{c}{\emph{$B$=50, $C$=10}} && \multicolumn{3}{c}{\emph{$B$=50, $C$=5}} && \multicolumn{3}{c}{\emph{$B$=10, $C$=10}} && \multicolumn{3}{c}{\emph{$B$=5, $C$=5}} \\
  \cmidrule{2-4} \cmidrule{6-8} \cmidrule{10-12} \cmidrule{14-16}
   & Avg ($\uparrow$) & Last ($\uparrow$) & $\mathcal{F}$ ($\downarrow$) && Avg ($\uparrow$) & Last ($\uparrow$) & $\mathcal{F}$ ($\downarrow$) &&  Avg ($\uparrow$) & Last ($\uparrow$) & $\mathcal{F}$ ($\downarrow$) && Avg ($\uparrow$) & Last ($\uparrow$) & $\mathcal{F}$ ($\downarrow$)\\
    \midrule
    Oracle  & 80.6 & 80.6 & - && 80.6 & 80.6 & - && 80.6 & 80.6 & - && 80.6 & 80.6 & - \\
    \quad w/ LingoCL & 80.6 & 80.6 & - && 80.6 & 80.6 & - && 80.6 & 80.6 & - && 80.6 & 80.6 & - \\
    \midrule
    \multicolumn{5}{l}{\textit{Architecture-based methods}} \\
    AANet \cite{aanet} & 75.3\scriptsize$\pm1.0$ & 66.2\scriptsize$\pm0.7$ & - && 72.7\scriptsize$\pm0.7$ & 61.2\scriptsize$\pm1.3$ & - && 57.8\scriptsize$\pm0.0$ & 42.9\scriptsize$\pm1.2$ & - && 48.5\scriptsize$\pm1.1$ & 37.9\scriptsize$\pm1.3$ & - \\
    \quad w/ LingoCL & \textbf{75.7}\scriptsize$\pm0.6$ & \textbf{68.7}\scriptsize$\pm0.5$ & - && \textbf{72.9}\scriptsize$\pm1.1$ & \textbf{61.7}\scriptsize$\pm0.9$ & - && \textbf{62.3}\scriptsize$\pm0.4$ & \textbf{46.9}\scriptsize$\pm0.3$ & - && \textbf{50.1}\scriptsize$\pm1.4$ & \textbf{41.3}\scriptsize$\pm0.7$ & - \\
    & \gr{+0.4} & \gr{+2.5} & - && \gr{+0.2} & \gr{+0.5} & - && \gr{+4.5} & \gr{+4.0} & - && \gr{+1.6} & \gr{+3.4} & - \\ 
    
    DyTox \cite{dytox} & 79.8\scriptsize$\pm0.4$ & 72.5\scriptsize$\pm1.1$ & 10.8\scriptsize$\pm0.3$ && 75.5\scriptsize$\pm1.1$ & 65.4\scriptsize$\pm1.3$ & 15.3\scriptsize$\pm1.4$ && 78.1\scriptsize$\pm0.0$ & 66.9\scriptsize$\pm0.8$ & 15.6\scriptsize$\pm0.8$ && 75.4\scriptsize$\pm0.9$ & 61.3\scriptsize$\pm1.3$ & 22.9\scriptsize$\pm0.3$ \\ 
    \quad w/ LingoCL & \textbf{80.6}\scriptsize$\pm0.2$ & \textbf{72.8}\scriptsize$\pm0.2$ & \textbf{6.9}\scriptsize$\pm1.2$ && \textbf{76.7}\scriptsize$\pm0.4$ & \textbf{66.8}\scriptsize$\pm0.6$ &  \textbf{12.3}\scriptsize$\pm0.2$ && \textbf{79.5}\scriptsize$\pm1.0$ & \textbf{69.6}\scriptsize$\pm0.9$ & \textbf{12.3}\scriptsize$\pm1.0$ && \textbf{76.2}\scriptsize$\pm0.3$ & \textbf{63.0}\scriptsize$\pm1.4$ & \textbf{18.3}\scriptsize$\pm0.5$ \\
    & \gr{+0.8} & \gr{+0.3} & \gr{-3.9} && \gr{+1.2} & \gr{+1.4} & \gr{-3.0} && \gr{+1.4} & \gr{+2.7} & \gr{-3.3} && \gr{+0.8} & \gr{+1.7} & \gr{-4.6} \\

    BiMeCo \cite{bimeco} & 71.2\scriptsize$\pm0.2$ & 60.7\scriptsize$\pm0.5$ & 15.8\scriptsize$\pm0.3$ && 68.9\scriptsize$\pm0.1$ & 59.7\scriptsize$\pm0.1$ & 22.5\scriptsize$\pm0.4$ && 59.0\scriptsize$\pm0.1$ & 44.0\scriptsize$\pm0.2$ & 41.1\scriptsize$\pm0.2$ && 47.6\scriptsize$\pm0.3$ & 35.5\scriptsize$\pm0.5$ & 49.0\scriptsize$\pm0.6$ \\ 
    \quad w/ LingoCL & \textbf{73.0}\scriptsize$\pm0.1$ & \textbf{63.4}\scriptsize$\pm0.3$ & \textbf{11.8}\scriptsize$\pm0.3$ && \textbf{71.1}\scriptsize$\pm0.2$ & \textbf{63.1}\scriptsize$\pm0.2$ &  \textbf{18.4}\scriptsize$\pm0.3$ && \textbf{60.0}\scriptsize$\pm0.7$ & \textbf{47.0}\scriptsize$\pm0.5$ & \textbf{38.6}\scriptsize$\pm0.2$ && \textbf{49.8}\scriptsize$\pm0.1$ & \textbf{39.4}\scriptsize$\pm0.2$ & \textbf{45.1}\scriptsize$\pm0.2$ \\
    & \gr{+1.8} & \gr{+2.7} & \gr{-4.0} && \gr{+2.2} & \gr{+3.4} & \gr{-4.1} && \gr{+1.0} & \gr{+3.0} & \gr{-2.5} && \gr{+2.2} & \gr{+3.9} & \gr{-3.9} \\
    
    \midrule
    \multicolumn{5}{l}{\textit{Distillation-based methods}} \\
    LUCIR \cite{lucir} & 70.2\scriptsize$\pm0.0$ & 59.7\scriptsize$\pm1.0$ & 21.4\scriptsize$\pm0.7$  &&  67.7\scriptsize$\pm0.8$ & 56.7\scriptsize$\pm0.2$  & 23.3\scriptsize$\pm1.4$  && 57.1\scriptsize$\pm0.8$ & 42.2\scriptsize$\pm0.4$ & 44.5\scriptsize$\pm0.7$ && 47.5\scriptsize$\pm0.5$ & 35.5\scriptsize$\pm1.0$ & 48.5\scriptsize$\pm1.5$ \\
    \quad w/ LingoCL & \textbf{73.4}\scriptsize$\pm1.1$ & \textbf{66.0}\scriptsize$\pm1.3$ & \textbf{8.3}\scriptsize$\pm0.9$  && \textbf{71.5}\scriptsize$\pm0.6$ & \textbf{62.3}\scriptsize$\pm1.3$  & \textbf{10.3}\scriptsize$\pm1.0$   && \textbf{62.2}\scriptsize$\pm0.9$ & \textbf{48.7}\scriptsize$\pm1.4$ & \textbf{38.6}\scriptsize$\pm0.7$ && \textbf{53.6}\scriptsize$\pm1.0$ & \textbf{41.9}\scriptsize$\pm1.4$ & \textbf{45.9}\scriptsize$\pm0.8$ \\
    & \gr{+3.2} & \gr{+6.3} & \gr{-13.1} && \gr{+3.8} & \gr{+5.6} & \gr{-13.0} && \gr{+5.1} & \gr{+6.5} & \gr{-5.9} && \gr{+6.1} & \gr{+6.4} & \gr{-2.6} \\
    
    BiC \cite{bic} & 72.0\scriptsize$\pm0.7$ & 63.6\scriptsize$\pm0.3$ & 7.2\scriptsize$\pm1.4$ && 68.1\scriptsize$\pm1.4$ & 57.5\scriptsize$\pm0.9$ & 7.1\scriptsize$\pm0.8$ && 66.4\scriptsize$\pm1.0$ & 55.2\scriptsize$\pm1.0$ & 15.1\scriptsize$\pm0.8$ && 59.0\scriptsize$\pm0.5$ & 44.8\scriptsize$\pm0.7$ & 20.9\scriptsize$\pm0.7$ \\
    \quad w/ LingoCL & \textbf{73.8}\scriptsize$\pm0.9$ & \textbf{65.8}\scriptsize$\pm0.1$ & \textbf{6.3}\scriptsize$\pm1.0$ && \textbf{72.8}\scriptsize$\pm0.4$ & \textbf{63.9}\scriptsize$\pm0.4$ & \textbf{4.1}\scriptsize$\pm0.4$ && \textbf{68.9}\scriptsize$\pm0.5$ & \textbf{58.4}\scriptsize$\pm1.5$ & \textbf{14.3}\scriptsize$\pm1.0$ && \textbf{61.9}\scriptsize$\pm0.1$ & \textbf{49.6}\scriptsize$\pm0.4$ & \textbf{18.8}\scriptsize$\pm1.2$ \\
    & \gr{+1.8} & \gr{+2.2} & \gr{-0.9} && \gr{+4.7} & \gr{+6.4} & \gr{-3.0} && \gr{+2.5} & \gr{+3.2} & \gr{-0.8} && \gr{+2.9} & \gr{+4.8} & \gr{-2.1} \\
    
    \midrule
    \multicolumn{5}{l}{\textit{Rectification-based methods}} \\

    CwD \cite{cwd} & 72.0\scriptsize$\pm1.1$ & 62.2\scriptsize$\pm1.3$ & 21.4\scriptsize$\pm0.3$ && 69.2\scriptsize$\pm1.3$ & 59.5\scriptsize$\pm0.5$ & 24.6\scriptsize$\pm0.8$ && 53.6\scriptsize$\pm0.0$ & 38.3\scriptsize$\pm0.6$ & 46.4\scriptsize$\pm0.7$ && 38.9\scriptsize$\pm0.4$ & 26.8\scriptsize$\pm1.1$ & 55.0\scriptsize$\pm1.1$\\
    \quad w/ LingoCL &  \textbf{73.4}\scriptsize$\pm1.1$ & \textbf{65.6}\scriptsize$\pm0.8$ & \textbf{10.9}\scriptsize$\pm1.2$ && \textbf{71.8}\scriptsize$\pm0.3$ & \textbf{62.8}\scriptsize$\pm1.3$ & \textbf{15.6}\scriptsize$\pm0.2$ && \textbf{55.6}\scriptsize$\pm1.3$ & \textbf{41.5}\scriptsize$\pm1.0$ & \textbf{43.0}\scriptsize$\pm0.6$ && \textbf{41.0}\scriptsize$\pm1.3$ & \textbf{29.9}\scriptsize$\pm0.4$ & \textbf{53.0}\scriptsize$\pm0.5$ \\ 
    & \gr{+1.4} & \gr{+3.4} & \gr{-10.5} && \gr{+2.6} & \gr{+3.3} & \gr{-9.0} && \gr{+2.0} & \gr{+3.2} & \gr{-3.4} && \gr{+2.1} & \gr{+3.1} & \gr{-2.0} \\
    
    IL2M \cite{il2m} & 67.5\scriptsize$\pm1.4$ & 54.6\scriptsize$\pm0.7$ & 30.0\scriptsize$\pm0.4$ && 63.6\scriptsize$\pm0.8$ & 50.8\scriptsize$\pm0.9$ & 33.7\scriptsize$\pm0.0$ && 55.0\scriptsize$\pm1.4$ & 39.9\scriptsize$\pm0.1$ & 50.2\scriptsize$\pm0.1$ && 46.4\scriptsize$\pm0.3$ & 34.6\scriptsize$\pm0.0$ & 50.1\scriptsize$\pm0.0$\\ 
    \quad w/ LingoCL & \textbf{71.7}\scriptsize$\pm0.6$ & \textbf{60.9}\scriptsize$\pm0.9$ & \textbf{23.8}\scriptsize$\pm1.2$ && \textbf{69.1}\scriptsize$\pm0.5$ & \textbf{57.4}\scriptsize$\pm1.3$ & \textbf{26.6}\scriptsize$\pm1.3$ && \textbf{59.9}\scriptsize$\pm1.3$ & \textbf{46.9}\scriptsize$\pm0.3$ & \textbf{43.5}\scriptsize$\pm0.3$ && \textbf{51.7}\scriptsize$\pm0.4$ & \textbf{41.4}\scriptsize$\pm1.2$ & \textbf{48.0}\scriptsize$\pm0.5$ \\
    & \gr{+4.2} & \gr{+6.3} & \gr{-6.2} && \gr{+5.5} & \gr{+6.6} & \gr{-7.1} && \gr{+4.9} & \gr{+7.0} & \gr{-6.7} && \gr{+5.3} & \gr{+6.8} & \gr{-2.1} \\
    
  \bottomrule
\end{tabular}
}
\caption{Results on class-incremental experiments on ImageNet-100.}
\label{tab:i100}
\end{table*}

\subsection{Experimental Setup} 
\textbf{Continual learning protocols.} We evaluate LingoCL on four common CL protocols, including: \textit{class-incremental learning (CIL)}, \textit{general few-shot class-incremental learning}, \textit{task-incremental learning}, and \textit{domain incremental learning}.

\noindent\textbf{Datasets.} We use CIFAR100~\cite{cifar} for task-IL, both CIFAR100 and ImageNet-100~\cite{Rebuffi2017iCaRLIC} for class-IL, and OfficeHome~\cite{officehome} for domain-IL. More details about datasets are shown in the \emph{supplementary material.}

\noindent\textbf{Architecture.} We employ MobileNetV2 for BiMeCo~\cite{bimeco} and ResNet18~\cite{he2016deep} for other CNN-based methods. For ViT-based methods, such as DyTox~\cite{dytox}, we follow the original implementation and use ConViT~\cite{convit}. As for the pretrained language model, we utilize the text transformer in CLIP-B/32~\cite{clip} pretrained on WIT-400M~\cite{clip}. Results about more language models are explored in Tab.~\ref{tab:language}.

\noindent\textbf{Metrics.} Following~\cite{de2021continual,masana2022class}, LingoCL is extensively evaluated by three metrics: last-step accuracy (Last), average incremental accuracy (Avg), and forgetting rate ($\mathcal{F}$).

\noindent\textbf{Baselines.} We comprehensively evaluate the effectiveness of the proposed method on eleven baselines, spanning various continual learning approaches. These include distillation-based methods such as LUCIR~\cite{lucir} and BiC~\cite{bic}, architecture-based methods like DyTox~\cite{dytox} and AANet~\cite{aanet}, rehearsal-based methods such as GEM~\cite{gem}, regularization-based methods including EWC~\cite{ewc}, MAS~\cite{mas}, and SI~\cite{si}, and rectification-based methods like IL2M~\cite{il2m} and CwD~\cite{cwd}. To ensure a fair comparison, we implement all the baseline methods using their officially released code or the widely recognized CL library~\cite{cl_benchmark,facil} in the research community and keep their original hyperparameters unchanged.

\begin{figure*}[h]
     \hspace{-0.2cm}
     \begin{subfigure}{0.5\textwidth}
     \centering
        \includegraphics[width=1.0\textwidth]{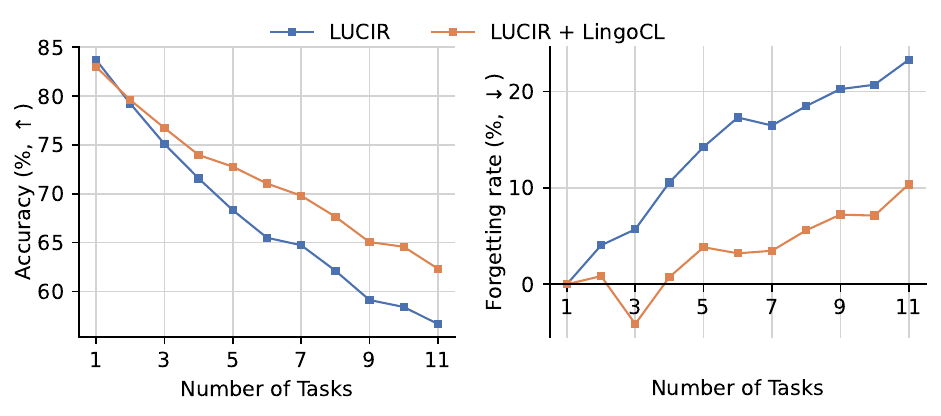}
        \caption{ImageNet-100 (B=50, C=5)}
	\label{fig:50_11}
    \end{subfigure}
    \begin{subfigure}{0.5\textwidth}
    \centering
        \includegraphics[width=1.0\textwidth]{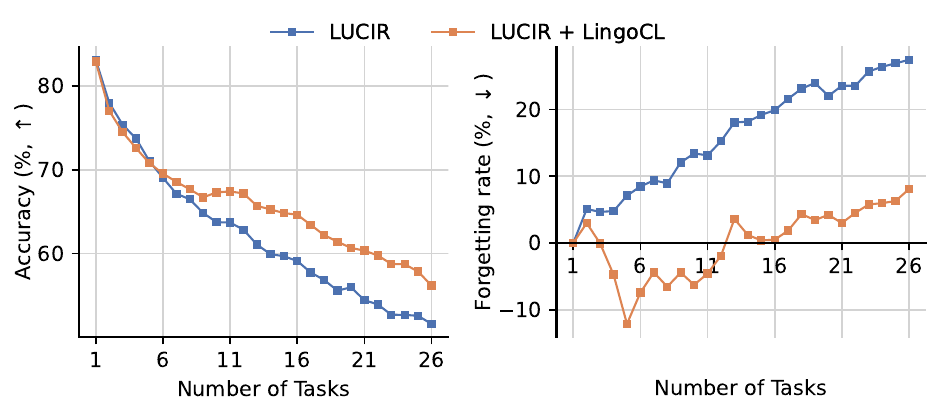}
        \caption{ImageNet-100 (B=50, C=2)}
	\label{fig:50_26}
    \end{subfigure}

    \caption{The evolution curve of accuracy and forgetting rate for each task on class-incremental experiments on ImageNet-100. Significantly, LingoCL exhibits negative forgetting, i.e., the learning of subsequent tasks leads to improved performance on prior tasks. This phenomenon evidences LingoCL's effective facilitation of knowledge transfer.} 
    \vspace{-0.3cm}
    \label{fig:lucir}
\end{figure*}
\begin{figure} 
\centering
    \includegraphics[width=0.5\textwidth]{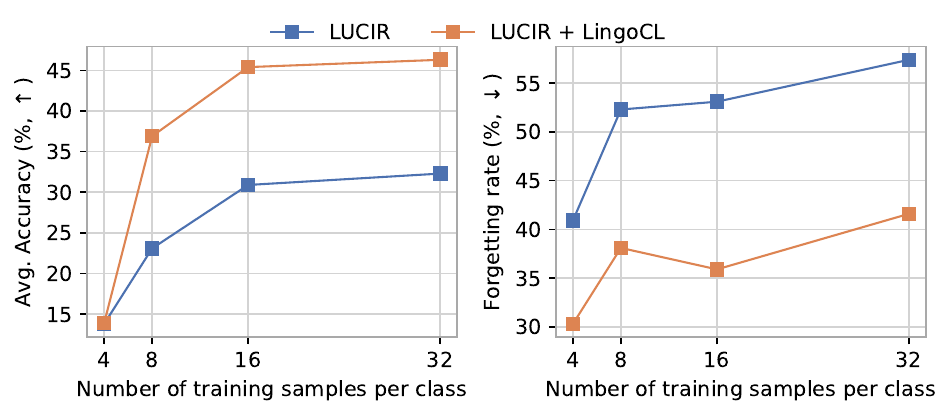}
    \caption{Results on general few-shot class-incremental learning.} 
    \vspace{-0.3cm}
    \label{fig:fewshot}
\end{figure}
\subsection{Class-incremental Learning Experiments} \label{sec:class_exp}
\noindent\textbf{Benchmark protocol.} We denote the number of classes in the initial task by $B$ and the number of new classes learned per task after the initial one by $C$. We adopt two popular protocols~\cite{Rebuffi2017iCaRLIC,bic}: 1) $B=50$: where the initial task covers half of the total number of classes and the remaining classes are equally divided among the subsequent tasks, and (2) $C=B$, where each task within the data stream involves an equal number of classes. The memory size for each class is set to 20 in all datasets. All approaches are evaluated under the same class order~\cite{Rebuffi2017iCaRLIC,aanet,lucir} for fair comparison.

\noindent\textbf{Implementation details.} We follow the original hyperparameters of all methods. We select exemplars as memory based on the herding strategy following previous works~\cite{Rebuffi2017iCaRLIC}. Detailed hyperparameters are in the \sm.

\noindent\textbf{Results.} We conduct extensive experiments by incorporating our method into various baselines of different routes. Tab.~\ref{tab:i100} and Tab.~\ref{tab:cifar} present the results on ImageNet-100 and CIFAR100, respectively, which illustrate that our method consistently and significantly improves all metrics. Taking LUCIR on ImageNet-100 as an example, our method improves the average accuracy by $3.2\% \sim 6.1\%$ across different settings. Importantly, our method barely impacts the performance of the oracle model, implying that the performance gains stem from reducing forgetting instead of the individual gains at each task. This is evidenced by the significant reductions in forgetting rate by $2.6\% \sim 13.1\%$. 

Moreover, we observe that the gains in last accuracy are usually larger than that in average accuracy, indicating that our method benefits more on long task sequences which are commonly more challenging. This observation is also supported by the experimental results. For instance, when the number of task increases from 6 to 11 (B=50,C=10 to B=50,C=5), the gains of CwD and LUCIR increase from $1.4\%$ and $3.2\%$ to $2.6\%$ and $3.8\%$, respectively.

Finally, in addition to the quantitative results, Fig.~\ref{fig:lucir} displays the accuracy and forgetting rate curves for LUCIR in long sequence settings. Our method achieves a smoother forgetting rate curve, with a gradual and consistent improvement in accuracy at each task. More importantly, our method even achieves negative forgetting rate, indicating that learning the later tasks helps to improve the accuracy of the previous ones. We attribute these gains to the fact that our method utilizes the semantic similarity among classes to guide the CL process, which promotes backward knowledge transfer.

\begin{table*}[t!]
\small
    \begin{subtable}[t]{0.33\textwidth}
    \centering
        \begin{tabular}{lcccc}
            \toprule
             Init. & Trainable & Avg ($\uparrow$) & Last ($\uparrow$) & $\mathcal{F}$ ($\downarrow$) \\
            \midrule
            \demph{Random}  & \demph{\cmark} & \demph{57.1\%} & \demph{42.2\%} & \demph{44.5\%} \\
            \midrule
            PLM & \cmark & 60.3\% & 47.8\% & 42.0\% \\
            \rowcolor[gray]{0.9} \textbf{PLM} & \xmark & \textbf{62.2\%} & \textbf{48.7\%} & \textbf{38.6\%} \\
            \bottomrule
        \end{tabular}
        \caption{Study on the effects of frozen weights.}
        \label{tab:frozen}
    \end{subtable}
    \hspace{0.25cm}
    \begin{subtable}[t]{0.36\textwidth}
    \centering
        \begin{tabular}{lccc}
            \toprule
             Corr. in weights & Avg ($\uparrow$) & Last ($\uparrow$) & $\mathcal{F}$ ($\downarrow$) \\
            \midrule
            \demph{Learned} & \demph{57.1\%} & \demph{42.2\%} & \demph{44.5\%} \\
            \midrule
            Orthogonal &  60.2\% & 47.5\% & 42.6\% \\
            \rowcolor[gray]{0.9} \textbf{PLM} & \textbf{62.2\%} & \textbf{48.7\%} & \textbf{38.6\%} \\
            \bottomrule
        \end{tabular}
        \caption{Study on the effects of semantic correlation.}
        \label{tab:sim}
    \end{subtable}
    \hspace{-0.1cm}
    \begin{subtable}[t]{0.31\textwidth}
    \centering
        \begin{tabular}{lccc}
            \toprule
             Sup. signal & Avg ($\uparrow$) & Last ($\uparrow$) & $\mathcal{F}$ ($\downarrow$) \\
            \midrule
            \demph{One-hot} & \demph{57.1\%} & \demph{42.2\%} & \demph{44.5\%} \\
            \midrule
            Oracle  & 61.6\% & \textbf{49.7\%} & 39.2\% \\
            \rowcolor[gray]{0.9} \textbf{PLM} & \textbf{62.2\%} & 48.7\% & \textbf{38.6\%} \\
            \bottomrule
        \end{tabular}
        \caption{Study on different supervision signal.}
        \label{tab:signal}
    \end{subtable}
\caption{Component analysis. The first cell  (\textcolor{demphcolor}{gray}) represents the baseline with a vanilla classifier which is trainable, randomly initialized, and supervised by one-hot label. Default settings are marked in \colorbox{baselinecolor}{gray}. ``Init.'': Initialization. ``Sup.": Supervision. ``Corr.": Correlation.}
\label{tab:component}
\vspace{-0.3cm}
\end{table*}

\begin{table}[t]
\center
\small
\resizebox{0.45\textwidth}{!}{
\begin{tabular}{lccccc}
\toprule 
\multirow{3}{*}{Method} & \multicolumn{2}{c}{Task-IL} && \multicolumn{2}{c}{Domain-IL} \\
\cmidrule{2-3} \cmidrule{5-6} 
& Last ($\uparrow$) & $\mathcal{F}$ ($\downarrow$) && Last ($\uparrow$) & $\mathcal{F}$ ($\downarrow$) \\
\midrule

Oracle & 77.6 & - && 99.1 & -   \\
\quad w/ LingoCL & 78.0 & - && 99.1 & -   \\
\midrule
\multicolumn{6}{l}{\textit{Regularization-based methods}} \\

EWC~\cite{ewc} & 41.5\scriptsize$\pm1.9$ & 45.2\scriptsize$\pm1.7$ && 48.4\scriptsize$\pm1.7$ & 45.9\scriptsize$\pm0.4$ \\
\quad w/ LingoCL  & \textbf{45.4}\scriptsize$\pm0.5$  &  \textbf{42.5}\scriptsize$\pm1.2$ && \textbf{52.2}\scriptsize$\pm0.9$ & \textbf{39.9}\scriptsize$\pm0.4$  \\
& \footnotesize\bgr{+3.9} &  \footnotesize\bgr{-2.7} && \footnotesize\bgr{+3.8} & \footnotesize\bgr{-6.0} \\

MAS~\cite{mas} & 42.9\scriptsize$\pm0.5$ & 44.5\scriptsize$\pm1.3$ && 54.1\scriptsize$\pm1.3$ & 41.9\scriptsize$\pm0.5$ \\
\quad w/ LingoCL & \textbf{47.0}\scriptsize$\pm1.5$ & \textbf{42.3}\scriptsize$\pm0.4$ && \textbf{58.1}\scriptsize$\pm2.0$ & \textbf{36.0}\scriptsize$\pm1.1$ \\
& \footnotesize\bgr{+4.1} & \footnotesize\bgr{-2.2} && \footnotesize\bgr{+4.0} & \footnotesize\bgr{-5.9}\\

GEM$^*$~\cite{gem} &- &- &&  58.5\scriptsize$\pm0.8$ & 38.4\scriptsize$\pm1.5$  \\
\quad w/ LingoCL & - & - && \textbf{59.7}\scriptsize$\pm0.5$  & \textbf{34.6}\scriptsize$\pm0.8$ \\
&  &  && \footnotesize\bgr{+1.2} & \footnotesize\bgr{-3.8} \\

SI~\cite{si} & 41.4\scriptsize$\pm1.7$ & 45.0\scriptsize$\pm1.2$ && 53.2\scriptsize$\pm1.8$ & 43.0\scriptsize$\pm1.1$ \\
\quad w/ LingoCL & \textbf{51.1}\scriptsize$\pm1.5$ & \textbf{35.4}\scriptsize$\pm0.8$ && \textbf{56.7}\scriptsize$\pm0.7$ & \textbf{35.8}\scriptsize$\pm1.9$ \\
&\footnotesize\bgr{+9.7} & \footnotesize\bgr{-9.6} && \footnotesize\bgr{+3.5} & \footnotesize\bgr{-7.2} \\

\bottomrule
\end{tabular}
}
\caption{Task-incremental and domain-incremental results. The method using an extra data buffer is marked with $^*$.}
\vspace{-0.2cm}
\label{table:task_domain_exp}
\end{table}

\subsection{General Few-shot Class-IL Experiments}
\noindent\textbf{Benchmark protocol.} General few-shot CIL is a more realistic setting where the initial task has sufficient training data to initialize the model, while the subsequent tasks only have $K$ training samples per class. No rehearsal buffer is available. In this study, we use ImageNet-100 as the benchmark dataset with $B=50$ and $C=10$ settings.

\noindent\textbf{Implementation details.} $K$ is set to 4/8/16/32 in our experiments. We choose LUCIR~\cite{lucir} as the baseline. 

\noindent\textbf{Results.} Due to the scarcity of data, few-shot CIL requires the model to not only overcome forgetting but also transfer as much learned knowledge as possible. As shown in Fig.~\ref{fig:fewshot}, our proposed method shows greater improvements in this challenging setting. Specifically, when $K$ is 32, our method achieves an improvement in accuracy of $14.0\%$ and a reduction in the forgetting rate of $15.8\%$. 
This demonstrates that our method achieves more effective knowledge transfer from the initial well-learned task by pre-allocating the semantic target for each class. We also observe that when $K$ is 4, although the gain of accuracy is marginal, the forgetting rate is reduced by $10.6\%$. It demonstrates that our method can alleviate the representation drifting in the feature space. See \textit{supplementary material} for more results.

\subsection{Task-incremental Learning Experiments} \label{sec:task_exp}
\noindent\textbf{Benchmark protocol.} The benchmark in this study is 10-split-CIFAR100, which involves dividing CIFAR100 into 10 tasks with non-overlapping classes. Following~\cite{null_space,l2p}, the last accuracy and forgetting rate are reported.

\noindent\textbf{Implementation details.} The learning rate is 1e-4 and epochs is 80. More detailed hyperparameters are shown in the \sm.

\noindent\textbf{Results.} As shown in Tab.~\ref{table:task_domain_exp}, our method improves the accuracy by $3.9\%\sim9.7\%$ and reduces the forgetting by $2.2\%\sim9.6\%$. Additionally, we observe that rehearsal-free methods can be competitive with rehearsal-based methods. SI~\cite{si} with our method achieves $51.1\%$ accuracy, surpassing the Rehearsal baseline by $3.0\%$. This finding suggests that our approach facilitates the effective use of learned knowledge, thus mitigating the reliance on old data.

\subsection{Domain-incremental Learning Experiments} \label{sec:domain_exp}
\noindent\textbf{Benchmark protocol.} OfficeHome~\cite{officehome} comprises four different domains, each treated as a distinct task. As the label set is consistent across all tasks, a shared classifier is utilized. We report the last accuracy and forgetting rate.

\noindent\textbf{Implementation details.} The regularization coefficients of EWC, MAS, SI and GEM are set to 100, 0.1, 0.3 and 5, respectively. More details are in the \sm.

\noindent\textbf{Results.} Tab.~\ref{table:task_domain_exp} reports that our method improves accuracy by $1.2\%\sim4.0\%$, while simultaneously reducing the forgetting rate by $3.8\%\sim7.6\%$. Due to the variability of the image domains, the semantic targets often shift or are limited to the current domains only. In contrast, the semantic targets generated by PLMs can utilize the rich source of domain knowledge in PLMs, ensuring a more representative distribution of these targets.

\subsection{Analysis and Ablation} \label{sec:ablation}
In this subsection, we conduct comprehensive ablation studies and analyses to systematically examine LingoCL. Unless stated otherwise, the experiments are based on LUCIR~\cite{lucir} and ImageNet-100 (B=10, C=10).

\noindent\textbf{Analysis of freezing the language-guided classifier.} We delve into two pivotal components in the design of the language-guided classifier: 1) freezing the weights, and 2) the semantic correlation in the weights. We first analyze the effect of freezing the weights in Tab.~\ref{tab:frozen}. The comparison between updating and freezing weights reveals that updates lead to a decrease in accuracy (from $61.7\%$ to $60.3\%$). This performance drop is attributed to catastrophic forgetting in semantic targets, triggered by updating weights for each task. It highlights the necessity of preserving the semantic knowledge sourced from pretrained language models. However, it's also notable that even with updated weights, performance exceeds that of random initialization, suggesting that strong initialization with rich semantics plays a crucial role in CL.

\noindent\textbf{Analysis of the semantic correlation in the classifier.} Furthermore, we ablate the semantic correlation in the classifier. By orthogonalizing the semantic targets output of the pretrained language model, we construct a classifier that removes semantic correlations among classes. The orthogonal classifier is kept frozen during training. As indicated in Tab.~\ref{tab:sim}, the removal of semantic correlation leads to a $2.0\%$ decrease in accuracy and a $4.0\%$ increase in the forgetting rate. Nevertheless, the orthogonal classifier still surpasses traditional vanilla classifiers by $3.1\%$ in accuracy. This suggests that the frozen, orthogonal targets help to reduce interference between different tasks, thereby diminishing feature drift in the feature space. On the other hand, the absence of semantic correlation appears to impede knowledge transfer across tasks. These findings underscore the dual significance of maintaining a frozen state and preserving semantic correlation in the classifier.

\noindent\textbf{Comparison with oracle classifier.} To thoroughly assess the impact of our language-guided supervision, we introduce an oracle classifier as a benchmark for oracle supervision. Initially, an idealized oracle model is trained with data from all tasks, typically considered the performance upper bound in CL. Subsequently, we replace the baseline model's vanilla classifier with this oracle classifier, which remains frozen during training. As shown in Tab.~\ref{tab:signal}, our language-guided classifier not only matches but surpasses the oracle classifier in both average accuracy and forgetting rate. This superiority is likely attributable to the fact that the dataset for pretraining language models is conceptually more diverse and sufficient than that used for the oracle model, providing semantically richer targets for each class. These results highlight the exceptional efficacy of our approach.

\begin{table}
  \centering
  \small
  \begin{tabular}{lcccc}
  \toprule
   \multirow{3}{*}{Method} & Forward & Semantic & \multirow{3}{*}{Avg ($\uparrow$)} & \multirow{3}{*}{$\mathcal{F}$ ($\downarrow$)} \\
   &  compatible & correlation & & \\
    \midrule
    Baseline & &  & 63.0\% & 29.9\% \\
    \quad w/ BiC~\cite{bic} &  \xmark & \xmark & 65.4\% & 25.1\% \\
    \quad w/ E2E~\cite{bic} &  \xmark & \xmark & 65.0\% & 26.1\% \\
    \quad w/ Div. head~\cite{bic} & \cmark & \xmark & 63.7\% & 27.2\% \\
    \rowcolor[gray]{0.9} \quad \textbf{w/ LingoCL} &  \cmark & \cmark & \textbf{67.5\%} & \textbf{22.5\%}  \\
  \bottomrule
\end{tabular}
\caption{Comparison with logits rectification-based methods.}
\label{tab:rectify}
\end{table}

\begin{table}[t!]
  \centering
  \resizebox{0.45\textwidth}{!}{
  \begin{tabular}{lcccc}
  \toprule
    Method & Pretraing data & Avg ($\uparrow$) & Last ($\uparrow$) & $\mathcal{F}$ ($\downarrow$) \\
    \midrule
    Baseline &  & 65.9\% & 55.8\% & 24.9\% \\
    \midrule
    \textit{multimodal pretraining} \\
    CLIP~\cite{clip} & WIT-400M & 67.5\% & 57.1\% & 22.5\% \\
    OpenCLIP~\cite{openclip} & LAION-2B & 68.0\% & 57.9\% & 22.0\% \\
    \midrule
    \textit{unimodal pretraining} \\
    BERT~\cite{bert} & 3.3B & 66.6\% & 57.1\% & 23.8\% \\
    XLNet~\cite{xlnet} & 32.8B & 67.0\% & 57.1\% & 22.8\% \\
  \bottomrule
\end{tabular}
}
\caption{Ablation study on the pretrained language models.}
\vspace{-0.5cm}
\label{tab:language}
\end{table}
\noindent\textbf{Comparison with logits rectification-based methods.} Tab.~\ref{tab:rectify} presents a comparison of LingoCL with other methods that modify the classifier to address anomalies. We use a simple CIL baseline with rehearsal and distillation on CIFAR100 under the setting of B=50, C=10. BiC~\cite{bic} addresses classifier bias by adding an extra linear layer, while EEIL~\cite{e2e} finetunes the classifier using balanced data. Divergence head~\cite{dytox} utilizes an additional classifier to separate the features of old and new tasks to preserve the feature space for future classes. Existing methods mainly focus on addressing the compatibility with old tasks using statistical corrections, whereas our method stands out by considering the semantic correlation among all classes, including the past and the future. Notably, LingoCL does not entirely conflict with these methods; in fact, LingoCL can complement it to further enhance performance, It is evidenced in Tab.~\ref{tab:cifar} and Tab.~\ref{tab:i100}, where LingoCL notably enhances the efficacy of BiC.

\noindent\textbf{Ablation on different language models.} In Tab.~\ref{tab:language}, we explore two types of language models: multimodal pretraining models and unimodal pretraining models. The overall results indicate that the multimodal pretraining language models perform better, which we attribute to the pretraining aligned with images allowing the language models to learn more semantic information from visual cues. Although the semantic targets generated by the unimodal pretraining models are not aligned with images, they still can be easily fitted with trainable vision encoders. Furthermore, we found that increasing the amount of pretraining data can effectively improve performance ($67.5\% \rightarrow 68.0\%$, $66.6\% \rightarrow 67.0\%$), as the language model learns more concepts.

\begin{figure}[t]
\centering
    \includegraphics[width=0.48\textwidth]{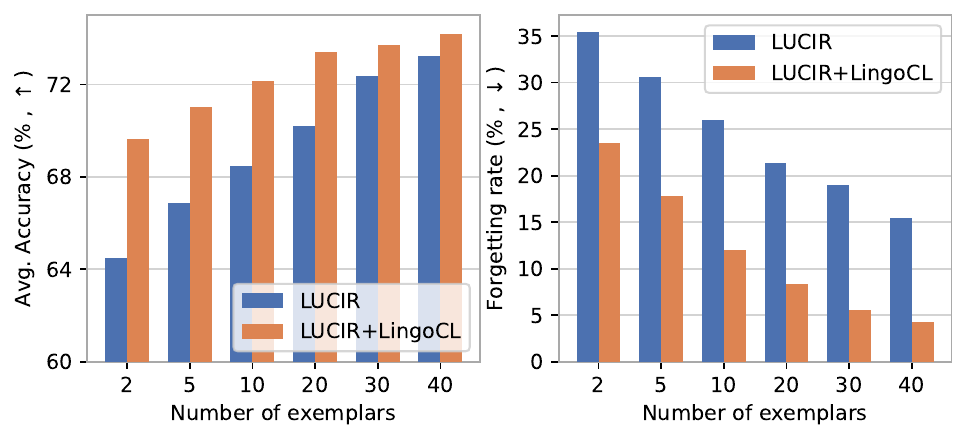}
    \caption{Effect of the number of exemplars for replay.}
    \label{fig:rehearsal}
\end{figure}
\begin{figure}[t]
\centering
    \includegraphics[width=0.5\textwidth]{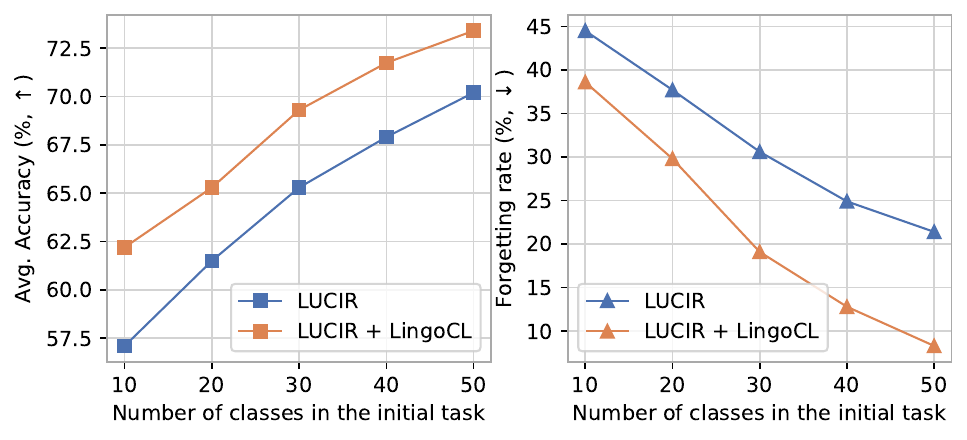}
    \caption{Impact of the number of classes in the initial task.}
    \label{fig:classes}
    \vspace{-0.5cm}
\end{figure}

\noindent\textbf{Effect of the number of exemplars for replay.} We investigate the effect of the number of old exemplars on model performance. The results in Fig.~\ref{fig:rehearsal} show that LingoCL can consistently improve the accuracy and reduce the forgetting rate under all settings, especially when the number of reserved exemplars is quite small. Notably, integrated with LingoCL, the baseline with reserving only 2 exemplars per class is comparable to the vanilla version by utilizing 20 exemplars per class ($69.6\%$ \textit{v.s.} $70.2\%$), further verifying the power of the proposed LingoCL.

\noindent\textbf{Impact of the number of classes in the initial task.} In this ablation, we discuss the effect of the number of classes learned in the initial task. As shown in Fig.~\ref{fig:classes}, the X-axis represents the number of classes in the initial task, and the remaining classes are incremented with 10 classes per task. We can observe that LingoCL can bring a $+3.2\%\sim5.1\%$ acreage accuracy improvement and reduce the forgetting rate by $+5.9\%\sim13.1\%$, which illustrates the effectiveness and robustness of our method.
\section{Conclusion}
In this work, we present a new perspective on CL, \ie, how to utilize the semantic knowledge in category names. Specifically, we use pretrained language models to generate the semantic target for each class. Empirical study shows that our method alleviates the representation drifting and facilitates knowledge transfer. Extensive experiments across various scenarios demonstrate the effectiveness of our method.

\noindent\textbf{Acknowledgements.} This work was supported partially by the National Natural Science Foundations of China (Grants No.62376267, 62076242), the Pre-Research Project on Civil Aerospace Technologies (No.D030312), the National Defense Basic Scientific Research Program of China(No.JCKY2021203B063) and the innoHK project.

{
    \small
    \bibliographystyle{ieeenat_fullname}
    \bibliography{main}

\begin{thebibliography}{59}
\providecommand{\natexlab}[1]{#1}
\providecommand{\url}[1]{\texttt{#1}}
\expandafter\ifx\csname urlstyle\endcsname\relax
  \providecommand{\doi}[1]{doi: #1}\else
  \providecommand{\doi}{doi: \begingroup \urlstyle{rm}\Url}\fi

\bibitem[Alayrac et~al.(2022)Alayrac, Donahue, Luc, Miech, Barr, Hasson, Lenc, Mensch, Millican, Reynolds, et~al.]{flamingo}
Jean-Baptiste Alayrac, Jeff Donahue, Pauline Luc, Antoine Miech, Iain Barr, Yana Hasson, Karel Lenc, Arthur Mensch, Katie Millican, Malcolm Reynolds, et~al.
\newblock Flamingo: a visual language model for few-shot learning.
\newblock In \emph{NeurIPS}, 2022.

\bibitem[Aljundi et~al.(2018)Aljundi, Babiloni, Elhoseiny, Rohrbach, and Tuytelaars]{mas}
Rahaf Aljundi, Francesca Babiloni, Mohamed Elhoseiny, Marcus Rohrbach, and Tinne Tuytelaars.
\newblock Memory aware synapses: Learning what (not) to forget.
\newblock In \emph{ECCV}, pages 139--154, 2018.

\bibitem[Aljundi et~al.(2019)Aljundi, Lin, Goujaud, and Bengio]{aljundi2019gradient}
Rahaf Aljundi, Min Lin, Baptiste Goujaud, and Yoshua Bengio.
\newblock Gradient based sample selection for online continual learning.
\newblock In \emph{NeurIPS}, 2019.

\bibitem[Belouadah and Popescu(2019)]{il2m}
Eden Belouadah and Adrian Popescu.
\newblock Il2m: Class incremental learning with dual memory.
\newblock In \emph{CVPR}, pages 583--592, 2019.

\bibitem[Brown et~al.(2020)Brown, Mann, Ryder, Subbiah, Kaplan, Dhariwal, Neelakantan, Shyam, Sastry, Askell, et~al.]{gpt3}
Tom Brown, Benjamin Mann, Nick Ryder, Melanie Subbiah, Jared~D Kaplan, Prafulla Dhariwal, Arvind Neelakantan, Pranav Shyam, Girish Sastry, Amanda Askell, et~al.
\newblock Language models are few-shot learners.
\newblock \emph{NeurIPS}, 33:\penalty0 1877--1901, 2020.

\bibitem[Castro et~al.(2018)Castro, Mar{\'\i}n-Jim{\'e}nez, Guil, Schmid, and Alahari]{e2e}
Francisco~M Castro, Manuel~J Mar{\'\i}n-Jim{\'e}nez, Nicol{\'a}s Guil, Cordelia Schmid, and Karteek Alahari.
\newblock End-to-end incremental learning.
\newblock In \emph{ECCV}, pages 233--248, 2018.

\bibitem[Chaudhry et~al.(2018)Chaudhry, Dokania, Ajanthan, and Torr]{rwalk}
Arslan Chaudhry, Puneet~K Dokania, Thalaiyasingam Ajanthan, and Philip~HS Torr.
\newblock Riemannian walk for incremental learning: Understanding forgetting and intransigence.
\newblock In \emph{ECCV}, pages 532--547, 2018.

\bibitem[Chen et~al.(2023)Chen, Huang, Chen, Geng, Zhang, Fang, Pan, and Chen]{chen2023duet}
Zhuo Chen, Yufeng Huang, Jiaoyan Chen, Yuxia Geng, Wen Zhang, Yin Fang, Jeff~Z Pan, and Huajun Chen.
\newblock Duet: Cross-modal semantic grounding for contrastive zero-shot learning.
\newblock In \emph{AAAI}, pages 405--413, 2023.

\bibitem[Cherti et~al.(2023)Cherti, Beaumont, Wightman, Wortsman, Ilharco, Gordon, Schuhmann, Schmidt, and Jitsev]{openclip}
Mehdi Cherti, Romain Beaumont, Ross Wightman, Mitchell Wortsman, Gabriel Ilharco, Cade Gordon, Christoph Schuhmann, Ludwig Schmidt, and Jenia Jitsev.
\newblock Reproducible scaling laws for contrastive language-image learning.
\newblock In \emph{CVPR}, 2023.

\bibitem[De~Lange et~al.(2021)De~Lange, Aljundi, Masana, Parisot, Jia, Leonardis, Slabaugh, and Tuytelaars]{de2021continual}
Matthias De~Lange, Rahaf Aljundi, Marc Masana, Sarah Parisot, Xu Jia, Ale{\v{s}} Leonardis, Gregory Slabaugh, and Tinne Tuytelaars.
\newblock A continual learning survey: Defying forgetting in classification tasks.
\newblock \emph{IEEE TPAMI}, 44\penalty0 (7):\penalty0 3366--3385, 2021.

\bibitem[Deng et~al.(2009)Deng, Dong, Socher, Li, Li, and Fei-Fei]{deng2009imagenet}
Jia Deng, Wei Dong, Richard Socher, Li-Jia Li, Kai Li, and Li Fei-Fei.
\newblock Imagenet: A large-scale hierarchical image database.
\newblock In \emph{CVPR}, 2009.

\bibitem[Devlin et~al.(2019)Devlin, Chang, Lee, and Toutanova]{bert}
Jacob Devlin, Ming-Wei Chang, Kenton Lee, and Kristina Toutanova.
\newblock Bert: Pre-training of deep bidirectional transformers for language understanding.
\newblock In \emph{NAACL}, 2019.

\bibitem[Douillard et~al.(2020)Douillard, Cord, Ollion, Robert, and Valle]{Douillard2020PODNetPO}
Arthur Douillard, Matthieu Cord, Charles Ollion, Thomas Robert, and Eduardo Valle.
\newblock Podnet: Pooled outputs distillation for small-tasks incremental learning.
\newblock In \emph{ECCV}, 2020.

\bibitem[Douillard et~al.(2022)Douillard, Ram{\'e}, Couairon, and Cord]{dytox}
Arthur Douillard, Alexandre Ram{\'e}, Guillaume Couairon, and Matthieu Cord.
\newblock Dytox: Transformers for continual learning with dynamic token expansion.
\newblock In \emph{CVPR}, 2022.

\bibitem[d’Ascoli et~al.(2021)d’Ascoli, Touvron, Leavitt, Morcos, Biroli, and Sagun]{convit}
St{\'e}phane d’Ascoli, Hugo Touvron, Matthew~L Leavitt, Ari~S Morcos, Giulio Biroli, and Levent Sagun.
\newblock Convit: Improving vision transformers with soft convolutional inductive biases.
\newblock In \emph{ICML}, pages 2286--2296. PMLR, 2021.

\bibitem[He et~al.(2016)He, Zhang, Ren, and Sun]{he2016deep}
Kaiming He, Xiangyu Zhang, Shaoqing Ren, and Jian Sun.
\newblock Deep residual learning for image recognition.
\newblock In \emph{CVPR}, 2016.

\bibitem[Hou et~al.(2019)Hou, Pan, Loy, Wang, and Lin]{lucir}
Saihui Hou, Xinyu Pan, Chen~Change Loy, Zilei Wang, and Dahua Lin.
\newblock Learning a unified classifier incrementally via rebalancing.
\newblock In \emph{CVPR}, pages 831--839, 2019.

\bibitem[Hsu et~al.(2018)Hsu, Liu, Ramasamy, and Kira]{cl_benchmark}
Yen-Chang Hsu, Yen-Cheng Liu, Anita Ramasamy, and Zsolt Kira.
\newblock Re-evaluating continual learning scenarios: A categorization and case for strong baselines.
\newblock In \emph{NeurIPS Workshop}, 2018.

\bibitem[Hu et~al.(2019)Hu, Lin, Liu, Tao, Tao, Zhao, Ma, and Yan]{hu2019overcoming}
Wenpeng Hu, Zhou Lin, Bing Liu, Chongyang Tao, Zhengwei~Tao Tao, Dongyan Zhao, Jinwen Ma, and Rui Yan.
\newblock Overcoming catastrophic forgetting for continual learning via model adaptation.
\newblock In \emph{ICLR}, 2019.

\bibitem[Jia et~al.(2021)Jia, Yang, Xia, Chen, Parekh, Pham, Le, Sung, Li, and Duerig]{align}
Chao Jia, Yinfei Yang, Ye Xia, Yi-Ting Chen, Zarana Parekh, Hieu Pham, Quoc Le, Yun-Hsuan Sung, Zhen Li, and Tom Duerig.
\newblock Scaling up visual and vision-language representation learning with noisy text supervision.
\newblock In \emph{ICML}, pages 4904--4916. PMLR, 2021.

\bibitem[Kemker and Kanan(2018)]{fearnet}
Ronald Kemker and Christopher Kanan.
\newblock Fearnet: Brain-inspired model for incremental learning.
\newblock In \emph{ICLR}, 2018.

\bibitem[Khan et~al.(2023)Khan, Naeem, Van~Gool, Stricker, Tombari, and Afzal]{khan2023introducing}
Muhammad Gul Zain~Ali Khan, Muhammad~Ferjad Naeem, Luc Van~Gool, Didier Stricker, Federico Tombari, and Muhammad~Zeshan Afzal.
\newblock Introducing language guidance in prompt-based continual learning.
\newblock In \emph{ICCV}, pages 11463--11473, 2023.

\bibitem[Kingma and Ba(2014)]{adam}
Diederik~P Kingma and Jimmy Ba.
\newblock Adam: A method for stochastic optimization.
\newblock \emph{arXiv preprint arXiv:1412.6980}, 2014.

\bibitem[Kirkpatrick et~al.(2017)Kirkpatrick, Pascanu, Rabinowitz, Veness, Desjardins, Rusu, Milan, Quan, Ramalho, Grabska-Barwinska, et~al.]{ewc}
James Kirkpatrick, Razvan Pascanu, Neil Rabinowitz, Joel Veness, Guillaume Desjardins, Andrei~A Rusu, Kieran Milan, John Quan, Tiago Ramalho, Agnieszka Grabska-Barwinska, et~al.
\newblock Overcoming catastrophic forgetting in neural networks.
\newblock \emph{PNAS}, 114\penalty0 (13):\penalty0 3521--3526, 2017.

\bibitem[Krizhevsky et~al.(2009)]{cifar}
Alex Krizhevsky et~al.
\newblock Learning multiple layers of features from tiny images.
\newblock 2009.

\bibitem[Li et~al.(2023)Li, Li, Savarese, and Hoi]{blip2}
Junnan Li, Dongxu Li, Silvio Savarese, and Steven Hoi.
\newblock Blip-2: Bootstrapping language-image pre-training with frozen image encoders and large language models.
\newblock \emph{arXiv preprint arXiv:2301.12597}, 2023.

\bibitem[Li et~al.(2022)Li, Xu, Dong, Zheng, Liu, Kong, and Sun]{vision_bring_to}
Lei Li, Jingjing Xu, Qingxiu~Dong Dong, Ce Zheng, Qi Liu, Lingpeng Kong, and Xu Sun.
\newblock What does vision supervision bring to language models? a case study of clip.
\newblock \emph{OpenReview}, 2022.

\bibitem[Li et~al.(2019)Li, Zhou, Wu, Socher, and Xiong]{learntogrow}
Xilai Li, Yingbo Zhou, Tianfu Wu, Richard Socher, and Caiming Xiong.
\newblock Learn to grow: A continual structure learning framework for overcoming catastrophic forgetting.
\newblock In \emph{ICML}, pages 3925--3934. PMLR, 2019.

\bibitem[Liu et~al.(2021)Liu, Schiele, and Sun]{aanet}
Yaoyao Liu, Bernt Schiele, and Qianru Sun.
\newblock Adaptive aggregation networks for class-incremental learning.
\newblock In \emph{CVPR}, pages 2544--2553, 2021.

\bibitem[Lopez-Paz and Ranzato(2017)]{gem}
David Lopez-Paz and Marc'Aurelio Ranzato.
\newblock Gradient episodic memory for continual learning.
\newblock In \emph{NeurIPS}, 2017.

\bibitem[Mallya and Lazebnik(2018)]{packnet}
Arun Mallya and Svetlana Lazebnik.
\newblock Packnet: Adding multiple tasks to a single network by iterative pruning.
\newblock In \emph{CVPR}, pages 7765--7773, 2018.

\bibitem[Mallya et~al.(2018)Mallya, Davis, and Lazebnik]{piggyback}
Arun Mallya, Dillon Davis, and Svetlana Lazebnik.
\newblock Piggyback: Adapting a single network to multiple tasks by learning to mask weights.
\newblock In \emph{ECCV}, pages 67--82, 2018.

\bibitem[Masana et~al.(2022{\natexlab{a}})Masana, Liu, Twardowski, Menta, Bagdanov, and van~de Weijer]{facil}
Marc Masana, Xialei Liu, Bartlomiej Twardowski, Mikel Menta, Andrew~D Bagdanov, and Joost van~de Weijer.
\newblock Class-incremental learning: Survey and performance evaluation on image classification.
\newblock \emph{IEEE TPAMI}, pages 1--20, 2022{\natexlab{a}}.

\bibitem[Masana et~al.(2022{\natexlab{b}})Masana, Liu, Twardowski, Menta, Bagdanov, and Van De~Weijer]{masana2022class}
Marc Masana, Xialei Liu, Bart{\l}omiej Twardowski, Mikel Menta, Andrew~D Bagdanov, and Joost Van De~Weijer.
\newblock Class-incremental learning: survey and performance evaluation on image classification.
\newblock \emph{IEEE TPAMI}, 45\penalty0 (5):\penalty0 5513--5533, 2022{\natexlab{b}}.

\bibitem[Ni et~al.(2022)Ni, Peng, Chen, Zhang, Meng, Fu, Xiang, and Ling]{ni2022expanding}
Bolin Ni, Houwen Peng, Minghao Chen, Songyang Zhang, Gaofeng Meng, Jianlong Fu, Shiming Xiang, and Haibin Ling.
\newblock Expanding language-image pretrained models for general video recognition.
\newblock In \emph{European Conference on Computer Vision}, pages 1--18. Springer, 2022.

\bibitem[Nie et~al.(2023)Nie, Xu, Liu, Meng, Huo, and Xiang]{bimeco}
Xing Nie, Shixiong Xu, Xiyan Liu, Gaofeng Meng, Chunlei Huo, and Shiming Xiang.
\newblock Bilateral memory consolidation for continual learning.
\newblock In \emph{CVPR}, pages 16026--16035, 2023.

\bibitem[Ostapenko et~al.(2019)Ostapenko, Puscas, Klein, Jahnichen, and Nabi]{ostapenko2019learning}
Oleksiy Ostapenko, Mihai Puscas, Tassilo Klein, Patrick Jahnichen, and Moin Nabi.
\newblock Learning to remember: A synaptic plasticity driven framework for continual learning.
\newblock In \emph{CVPR}, pages 11321--11329, 2019.

\bibitem[Prabhu et~al.(2020)Prabhu, Torr, and Dokania]{gdumb}
Ameya Prabhu, Philip~HS Torr, and Puneet~K Dokania.
\newblock Gdumb: A simple approach that questions our progress in continual learning.
\newblock In \emph{ECCV}, pages 524--540, 2020.

\bibitem[Radford et~al.(2021)Radford, Kim, Hallacy, Ramesh, Goh, Agarwal, Sastry, Askell, Mishkin, Clark, et~al.]{clip}
Alec Radford, Jong~Wook Kim, Chris Hallacy, Aditya Ramesh, Gabriel Goh, Sandhini Agarwal, Girish Sastry, Amanda Askell, Pamela Mishkin, Jack Clark, et~al.
\newblock Learning transferable visual models from natural language supervision.
\newblock In \emph{ICML}, pages 8748--8763. PMLR, 2021.

\bibitem[Ramasesh et~al.(2021)Ramasesh, Dyer, and Raghu]{anatomy}
Vinay~V Ramasesh, Ethan Dyer, and Maithra Raghu.
\newblock Anatomy of catastrophic forgetting: Hidden representations and task semantics.
\newblock In \emph{ICLR}, 2021.

\bibitem[Rebuffi et~al.(2017)Rebuffi, Kolesnikov, Sperl, and Lampert]{Rebuffi2017iCaRLIC}
Sylvestre-Alvise Rebuffi, Alexander Kolesnikov, G. Sperl, and Christoph~H. Lampert.
\newblock icarl: Incremental classifier and representation learning.
\newblock In \emph{CVPR}, pages 5533--5542, 2017.

\bibitem[Robbins and Monro(1951)]{sgd}
Herbert Robbins and Sutton Monro.
\newblock A stochastic approximation method.
\newblock \emph{The annals of mathematical statistics}, pages 400--407, 1951.

\bibitem[Rusu et~al.(2016)Rusu, Rabinowitz, Desjardins, Soyer, Kirkpatrick, Kavukcuoglu, Pascanu, and Hadsell]{pnn}
Andrei~A Rusu, Neil~C Rabinowitz, Guillaume Desjardins, Hubert Soyer, James Kirkpatrick, Koray Kavukcuoglu, Razvan Pascanu, and Raia Hadsell.
\newblock Progressive neural networks.
\newblock \emph{arXiv preprint arXiv:1606.04671}, 2016.

\bibitem[Shi et~al.(2022)Shi, Zhou, Liang, Jiang, Feng, Torr, Bai, and Tan]{cwd}
Yujun Shi, Kuangqi Zhou, Jian Liang, Zihang Jiang, Jiashi Feng, Philip~HS Torr, Song Bai, and Vincent~YF Tan.
\newblock Mimicking the oracle: an initial phase decorrelation approach for class incremental learning.
\newblock In \emph{CVPR}, pages 16722--16731, 2022.

\bibitem[Tao et~al.(2020{\natexlab{a}})Tao, Chang, Hong, Wei, and Gong]{tao2020topology}
Xiaoyu Tao, Xinyuan Chang, Xiaopeng Hong, Xing Wei, and Yihong Gong.
\newblock Topology-preserving class-incremental learning.
\newblock In \emph{ECCV}, pages 254--270, 2020{\natexlab{a}}.

\bibitem[Tao et~al.(2020{\natexlab{b}})Tao, Hong, Chang, Dong, Wei, and Gong]{tao2020few}
Xiaoyu Tao, Xiaopeng Hong, Xinyuan Chang, Songlin Dong, Xing Wei, and Yihong Gong.
\newblock Few-shot class-incremental learning.
\newblock In \emph{CVPR}, pages 12183--12192, 2020{\natexlab{b}}.

\bibitem[Tsimpoukelli et~al.(2021)Tsimpoukelli, Menick, Cabi, Eslami, Vinyals, and Hill]{frozen}
Maria Tsimpoukelli, Jacob~L Menick, Serkan Cabi, SM Eslami, Oriol Vinyals, and Felix Hill.
\newblock Multimodal few-shot learning with frozen language models.
\newblock In \emph{NeurIPS}, pages 200--212, 2021.

\bibitem[Venkateswara et~al.(2017)Venkateswara, Eusebio, Chakraborty, and Panchanathan]{officehome}
Hemanth Venkateswara, Jose Eusebio, Shayok Chakraborty, and Sethuraman Panchanathan.
\newblock Deep hashing network for unsupervised domain adaptation.
\newblock In \emph{CVPR}, pages 5018--5027, 2017.

\bibitem[Wang et~al.(2023)Wang, Xu, Hu, Yan, Sang, and Qian]{wang2023improved}
Junyang Wang, Yuanhong Xu, Juhua Hu, Ming Yan, Jitao Sang, and Qi Qian.
\newblock Improved visual fine-tuning with natural language supervision.
\newblock In \emph{ICCV}, 2023.

\bibitem[Wang et~al.(2021)Wang, Li, Sun, and Xu]{null_space}
Shipeng Wang, Xiaorong Li, Jian Sun, and Zongben Xu.
\newblock Training networks in null space of feature covariance for continual learning.
\newblock In \emph{CVPR}, pages 184--193, 2021.

\bibitem[Wang et~al.(2022)Wang, Zhang, Lee, Zhang, Sun, Ren, Su, Perot, Dy, and Pfister]{l2p}
Zifeng Wang, Zizhao Zhang, Chen-Yu Lee, Han Zhang, Ruoxi Sun, Xiaoqi Ren, Guolong Su, Vincent Perot, Jennifer Dy, and Tomas Pfister.
\newblock Learning to prompt for continual learning.
\newblock In \emph{CVPR}, pages 139--149, 2022.

\bibitem[Wu et~al.(2019)Wu, Chen, Wang, Ye, Liu, Guo, and Fu]{bic}
Yue Wu, Yinpeng Chen, Lijuan Wang, Yuancheng Ye, Zicheng Liu, Yandong Guo, and Yun Fu.
\newblock Large scale incremental learning.
\newblock In \emph{CVPR}, pages 374--382, 2019.

\bibitem[Yan et~al.(2021)Yan, Xie, and He]{der}
Shipeng Yan, Jiangwei Xie, and Xuming He.
\newblock Der: Dynamically expandable representation for class incremental learning.
\newblock In \emph{CVPR}, pages 3014--3023, 2021.

\bibitem[Yang et~al.(2019)Yang, Dai, Yang, Carbonell, Salakhutdinov, and Le]{xlnet}
Zhilin Yang, Zihang Dai, Yiming Yang, Jaime Carbonell, Russ~R Salakhutdinov, and Quoc~V Le.
\newblock Xlnet: Generalized autoregressive pretraining for language understanding.
\newblock \emph{NeurIPS}, 32, 2019.

\bibitem[Yuan et~al.(2021)Yuan, Chen, Chen, Codella, Dai, Gao, Hu, Huang, Li, Li, et~al.]{florence}
Lu Yuan, Dongdong Chen, Yi-Ling Chen, Noel Codella, Xiyang Dai, Jianfeng Gao, Houdong Hu, Xuedong Huang, Boxin Li, Chunyuan Li, et~al.
\newblock Florence: A new foundation model for computer vision.
\newblock \emph{arXiv preprint arXiv:2111.11432}, 2021.

\bibitem[Zenke et~al.(2017)Zenke, Poole, and Ganguli]{si}
Friedemann Zenke, Ben Poole, and Surya Ganguli.
\newblock Continual learning through synaptic intelligence.
\newblock In \emph{ICML}, pages 3987--3995, 2017.

\bibitem[Zhao et~al.(2020)Zhao, Xiao, Gan, Zhang, and Xia]{wa}
Bowen Zhao, Xi Xiao, Guojun Gan, Bin Zhang, and Shutao Xia.
\newblock Maintaining discrimination and fairness in class incremental learning.
\newblock In \emph{CVPR}, pages 13205--13214, 2020.

\bibitem[Zhao et~al.(2024)Zhao, Ni, Wang, Fan, Zhu, Wang, Chen, Meng, and Zhang]{zhao2024continual}
Hongbo Zhao, Bolin Ni, Haochen Wang, Junsong Fan, Fei Zhu, Yuxi Wang, Yuntao Chen, Gaofeng Meng, and Zhaoxiang Zhang.
\newblock Continual forgetting for pre-trained vision models.
\newblock \emph{arXiv preprint arXiv:2403.11530}, 2024.

\bibitem[Zhong et~al.(2018)Zhong, Yan, Wu, Shao, and Liu]{zhong2018practical}
Zhao Zhong, Junjie Yan, Wei Wu, Jing Shao, and Cheng-Lin Liu.
\newblock Practical block-wise neural network architecture generation.
\newblock In \emph{CVPR}, pages 2423--2432, 2018.

\end{thebibliography}
}

\clearpage
\setcounter{page}{1}
\maketitlesupplementary

In this supplementary material, we provide additional details regarding the main manuscript. More specifically:

\begin{itemize}
    \item In Sec.~\ref{sec:suppl_data_protocol_metric}, we provide further explanation of the datasets, protocols and metrics.
    \item In Sec.~\ref{sec:suppl_hyper}, we provide the detailed hyperparameters of different continual learning settings.
    \item In Sec.~\ref{sec:suppl_results}, we provide additional experiments and results.
\end{itemize}

\setlength{\tabcolsep}{1.0mm}{
\begin{table*}[t]
\centering
\resizebox{\textwidth}{!}{
\begin{tabular}
{p{5.5cm}p{1.8cm}p{1.8cm}p{1.6cm}p{0.2cm}p{1.8cm}p{1.8cm}p{1.8cm}p{0.2cm}p{1.8cm}p{1.8cm}p{1.7cm}}
\toprule
\multirow{3}{*}{Method} & \multicolumn{3}{c}{\emph{$B$=50, $C$=10}} && \multicolumn{3}{c}{\emph{$B$=50, $C$=5}} && \multicolumn{3}{c}{\emph{$B$=50, $C$=2}} \\
\cmidrule{2-4} \cmidrule{6-8} \cmidrule{10-12}
& Avg ($\uparrow$) & Last ($\uparrow$) & $\mathcal{F}$ ($\downarrow$) && Avg ($\uparrow$) & Last ($\uparrow$) & $\mathcal{F}$ ($\downarrow$) &&  Avg ($\uparrow$) & Last ($\uparrow$) & $\mathcal{F}$ ($\downarrow$) \\
\midrule
Oracle  & 77.6 & 77.6 & - && 77.6 & 77.6 & - && 77.6 & 77.6 & -  \\
\quad w/ LingoCL & 78.0 & 78.0 & - && 78.0 & 78.0 & - && 78.0 & 78.0 & - \\
\midrule
\multicolumn{5}{l}{\textit{Distillation-based methods}} \\
LUCIR~\cite{lucir} & 65.9\scriptsize$\pm1.7$ & 55.8\scriptsize$\pm1.7$ & 24.9\scriptsize$\pm1.7$ && 60.9\scriptsize$\pm1.1$ & 51.4\scriptsize$\pm0.4$ & 26.7\scriptsize$\pm1.5$ && 52.9\scriptsize$\pm0.5$ & 42.5\scriptsize$\pm1.5$ & 34.0\scriptsize$\pm0.6$ \\
\quad  w/ LingoCL & \textbf{67.5}\scriptsize$\pm0.8$ \gr{+1.6} & \textbf{57.1}\scriptsize$\pm1.5$ \gr{+1.3} & \textbf{22.5}\scriptsize$\pm0.6$ \gr{-2.4} && \textbf{62.1}\scriptsize$\pm1.9$ \gr{+1.2} & \textbf{53.2}\scriptsize$\pm2.0$ \gr{+1.8} & \textbf{22.4}\scriptsize$\pm1.6$ \gr{-4.3} && \textbf{53.5}\scriptsize$\pm1.2$ \gr{+0.6} & \textbf{43.4}\scriptsize$\pm1.6$ \gr{+0.9} & \textbf{31.2}\scriptsize$\pm0.5$ \gr{-2.8} \\

BiC~\cite{bic} & 63.5\scriptsize$\pm0.9$ & 51.2\scriptsize$\pm0.5$ & 16.2\scriptsize$\pm1.0$ && 57.0\scriptsize$\pm0.1$ & 45.0\scriptsize$\pm0.9$ & 15.9\scriptsize$\pm1.0$ && 44.6\scriptsize$\pm0.0$ & 33.2\scriptsize$\pm0.8$ & 11.1\scriptsize$\pm0.7$ \\ 
\quad w/ LingoCL & \textbf{64.7}\scriptsize$\pm0.9$ \gr{+1.2} & \textbf{52.4}\scriptsize$\pm0.8$ \gr{+1.2} & \textbf{11.8}\scriptsize$\pm1.0$ \gr{-4.4} && \textbf{58.6}\scriptsize$\pm0.5$ \gr{+1.6} & \textbf{46.3}\scriptsize$\pm0.9$ \gr{+1.3} & \textbf{14.5}\scriptsize$\pm0.8$ \gr{-1.4} && \textbf{46.7}\scriptsize$\pm0.6$ \gr{+2.1} & \textbf{35.1}\scriptsize$\pm0.9$ \gr{+1.9} & \textbf{6.3}\scriptsize$\pm0.8$ \gr{-4.8} \\

\midrule
\multicolumn{5}{l}{\textit{Rectification-based methods}} \\

CwD~\cite{cwd} & 66.9\scriptsize$\pm0.3$ & 57.4\scriptsize$\pm0.8$ & 23.4\scriptsize$\pm0.9$ && 62.3\scriptsize$\pm0.8$ & 52.5\scriptsize$\pm0.7$ & 26.9\scriptsize$\pm0.6$ && 56.3\scriptsize$\pm0.4$ & 44.7\scriptsize$\pm0.8$ & 36.2\scriptsize$\pm0.5$ \\
\quad w/ LingoCL & \textbf{68.0}\scriptsize$\pm0.4$ \gr{+1.1} & \textbf{58.4}\scriptsize$\pm0.2$ \gr{+1.0}& \textbf{22.8}\scriptsize$\pm0.9$ \gr{-0.6} && \textbf{63.3}\scriptsize$\pm1.0$ \gr{+1.0} & \textbf{53.6}\scriptsize$\pm0.5$ \gr{+1.1} & \textbf{26.0}\scriptsize$\pm0.5$ \gr{-0.9} && \textbf{58.3}\scriptsize$\pm0.2$ \gr{+2.0} & \textbf{46.1}\scriptsize$\pm0.5$ \gr{+1.4} & \textbf{34.1}\scriptsize$\pm0.3$ \gr{-2.1} \\

IL2M \cite{il2m} & 65.7\scriptsize$\pm0.1$ & 55.9\scriptsize$\pm0.3$ & 25.2\scriptsize$\pm0.7$ && 59.9\scriptsize$\pm0.6$ & 49.9\scriptsize$\pm0.1$ & 29.7\scriptsize$\pm0.2$ && 52.5\scriptsize$\pm0.8$ & 42.0\scriptsize$\pm0.3$ & 35.3\scriptsize$\pm0.6$ \\
\quad w/ LingoCL & \textbf{68.5}\scriptsize$\pm0.2$ \gr{+2.8} & \textbf{59.2}\scriptsize$\pm0.6$ \gr{+3.3} & \textbf{21.2}\scriptsize$\pm0.2$ \gr{-4.0} && \textbf{60.8}\scriptsize$\pm0.9$ \gr{+0.9} & \textbf{49.8}\scriptsize$\pm0.7$ \gr{-0.1} & \textbf{29.4}\scriptsize$\pm0.5$ \gr{-0.3} && \textbf{54.0}\scriptsize$\pm0.4$ \gr{+1.5} & \textbf{45.7}\scriptsize$\pm0.5$ \gr{+3.7} & \textbf{29.2}\scriptsize$\pm0.3$ \gr{-6.1} \\

\bottomrule
\end{tabular}
}
\caption{Results on class-incremental experiments on CIFAR100 of Average accuracy (\%), last phase accuracy (\%) and forgetting rate $\mathcal{F}$ (\%) with and without language-guided representation at various CL settings. $B$ denotes the number of classes at the initial task, and $C$ denotes the number of classes in each task after the initial one.}
\label{tab:suppl_cifar}

\end{table*}
}
\begin{table*}[t]
  \centering
  \resizebox{1\textwidth}{!}{
  \begin{tabular}{lccccccccccccccc}
  \toprule
   \multirow{3}{*}{Method} & \multicolumn{3}{c}{\emph{K=4}} && \multicolumn{3}{c}{\emph{K=8}} && \multicolumn{3}{c}{\emph{K=16}} && \multicolumn{3}{c}{\emph{K=32}} \\
  \cmidrule{2-4} \cmidrule{6-8} \cmidrule{10-12} \cmidrule{14-16}
   & Avg ($\uparrow$) & Last ($\uparrow$) & $\mathcal{F}$ ($\downarrow$) && Avg ($\uparrow$) & Last ($\uparrow$) & $\mathcal{F}$ ($\downarrow$) &&  Avg ($\uparrow$) & Last ($\uparrow$) & $\mathcal{F}$ ($\downarrow$) && Avg ($\uparrow$) & Last ($\uparrow$) & $\mathcal{F}$ ($\downarrow$)\\
    \midrule
    \multicolumn{5}{l}{\textit{Buffer size = 0}} \\
    LUCIR \cite{lucir} & 13.9\scriptsize$\pm0.4$ & 7.3\scriptsize$\pm0.6$ & 40.9\scriptsize$\pm0.4$ && 23.1\scriptsize$\pm0.8$ & 10.7\scriptsize$\pm1.0$ & 52.3\scriptsize$\pm0.6$ && 30.9\scriptsize$\pm0.3$ & 12.8\scriptsize$\pm0.1$ & 53.1\scriptsize$\pm0.1$ && 32.3\scriptsize$\pm0.3$ & 12.9\scriptsize$\pm0.8$ & 57.4\scriptsize$\pm0.8$ \\
    \quad w/ LingoCL & \textbf{13.9}\scriptsize$\pm0.2$ & \textbf{6.0}\scriptsize$\pm0.9$ & \textbf{30.2}\scriptsize$\pm0.5$ && \textbf{36.9}\scriptsize$\pm0.4$ & \textbf{12.7}\scriptsize$\pm0.5$ & \textbf{38.1}\scriptsize$\pm0.3$ && \textbf{45.4}\scriptsize$\pm0.8$ & \textbf{20.0}\scriptsize$\pm0.9$ & \textbf{35.9}\scriptsize$\pm0.2$ && \textbf{46.3}\scriptsize$\pm0.6$ & \textbf{20.4}\scriptsize$\pm0.5$ & \textbf{41.6}\scriptsize$\pm0.1$ \\
    & \gr{+0.0} & \gr{-1.3} & \gr{-10.7} && \gr{+13.8} & \gr{+2.0} & \gr{-14.2} && \gr{+14.5} & \gr{+7.2} & \gr{-17.2} && \gr{+14.0} & \gr{+7.5} & \gr{-15.8} \\
    
    \midrule
    
    \multicolumn{5}{l}{\textit{Buffer size = 1}} \\
    LUCIR~\cite{lucir} & 39.1\scriptsize$\pm0.2$ & 15.2\scriptsize$\pm0.3$ & 29.1\scriptsize$\pm0.9$ && 40.2\scriptsize$\pm0.1$ & 18.7\scriptsize$\pm0.5$ & 35.0\scriptsize$\pm0.4$ && 31.1\scriptsize$\pm0.8$ & 19.2\scriptsize$\pm0.4$ & 31.7\scriptsize$\pm0.4$ && 38.7\scriptsize$\pm0.9$ & 23.9\scriptsize$\pm0.0$ & 37.9\scriptsize$\pm0.3$ \\
    \quad w/ LingoCL & \textbf{41.6}\scriptsize$\pm0.2$ & \textbf{19.3}\scriptsize$\pm0.9$ & \textbf{10.6}\scriptsize$\pm0.2$ && \textbf{44.9}\scriptsize$\pm0.8$ & \textbf{21.8}\scriptsize$\pm0.8$ & \textbf{13.3}\scriptsize$\pm0.2$ && \textbf{38.9}\scriptsize$\pm0.3$ & \textbf{24.7}\scriptsize$\pm0.2$ & \textbf{13.8}\scriptsize$\pm0.5$ && \textbf{44.9}\scriptsize$\pm0.6$ & \textbf{29.2}\scriptsize$\pm0.3$ & \textbf{18.9}\scriptsize$\pm0.9$ \\
    & \gr{+2.5} & \gr{+4.1} & \gr{-18.5} && \gr{+4.7} & \gr{+3.1} & \gr{-21.7} && \gr{+7.8} & \gr{+5.5} & \gr{-17.9} && \gr{+6.2} & \gr{+5.3} & \gr{-19.0} \\
    
  \bottomrule
\end{tabular}
}
\caption{Results on few-shot class-incremental experiments on ImageNet100 under $B=50, C=10$.}
\label{tab:suppl_few}
\end{table*}

\section{Datasets, Protocols and Metrics}\label{sec:suppl_data_protocol_metric}
In Sec.~\ref{sec:suppl_dataset}, we present the statistical information for the datasets used in our experiments. In Sec.~\ref{sec:suppl_protocol}, we describe the continual learning protocols that are commonly used in the literature. Finally, in Sec.~\ref{sec:suppl_metric}, we introduce the evaluation metrics used to measure the performance comprehensively.

\subsection{Datasets statistics}\label{sec:suppl_dataset}

\vspace{1mm}
\begin{itemize}
\item {\em Split-CIFAR-100 (Task-IL)}. The CIFAR100 dataset~\cite{cifar} comprises 60,000 32×32 images belonging to 100 classes. In task-incremental learning setting, Split-CIFAR-100 splits the original CIFAR-100~\cite{cifar} into 10 tasks, 10 disjoint classes per task. 
\end{itemize}

\vspace{1mm}
\begin{itemize}
\item {\em CIFAR-100 (Class-IL)}. In the class-incremental learning setting, we divide the classes into mutually exclusive sets. The first task consists of $B$ classes, and each subsequent task consists of $C$ classes.
\end{itemize}

\vspace{1mm}
\begin{itemize}
\item {\em ImageNet-100 (Class-IL)}. ImageNet-100~\cite{Rebuffi2017iCaRLIC} is the subset of ImageNet1000~\cite{deng2009imagenet} containing 100 classes~\cite{Rebuffi2017iCaRLIC}. These classes are selected from the first 100 classes after a random shuffle with seed 1,993~\cite{zhong2018practical}. Each image is represented by 224×224 pixels.
\end{itemize}

\vspace{1mm}
\begin{itemize}
\item {\em OfficeHome (Domain-IL)}. OfficeHome~\cite{officehome} consists of images from four different domains: Artistic images, Clip Art, Product images and Real-World images. For each domain, the dataset contains images of 65 object categories found typically in Office and Home settings. Each image is represented by 224×224 pixels.
\end{itemize}

We use the official categories provided by the respective dataset creators for all datasets, which can be accessed through the dataset resources~\cite{cifar,deng2009imagenet,officehome}. These categories are also presented in Fig.~\ref{fig:suppl_cate_cifar}, Fig.~\ref{fig:suppl_cate_i100}, and Fig.~\ref{fig:suppl_cate_office} for CIFAR100, ImageNet100, and OfficeHome, respectively.

\subsection{Continual Learning Protocols}\label{sec:suppl_protocol}
In continual learning (CL), the model is trained in a task-by-task manner. We define a sequence of tasks denoted by $\mathcal{D} = \{\mathcal{D}_1, \cdots, \mathcal{D}_T\}$. The $t$-th task, denoted by $\mathcal{D}_t = \{(\boldsymbol{x}^t_i, y^t_i)\}_{i=1}^{n_t}$, comprises tuples consisting of an input sample $\boldsymbol{x}^t \in \mathcal{X}_t$ and its corresponding label $y^t \in \mathcal{Y}_t$. Depending on the target set and the number of training samples, CL protocols can be divided into four common categories:

\begin{itemize}
\item {\em Task-incremental learning} where the target set of test sample $\boldsymbol{x}^t$ is $\mathcal{Y}_t$.
\end{itemize}

\begin{itemize}
\item {\em Class-incremental learning} where the target set of test sample $\boldsymbol{x}^t$ is $\cup_{i=1}^{t}\mathcal{Y}_i$.
\end{itemize}

\begin{itemize}
\item {\em Few-shot Class-incremental learning} where the target set of test sample $\boldsymbol{x}^t$ is $\cup_{i=1}^{t}\mathcal{Y}_i$ and the $n_t (t > 1)$ of training set is limited.
\end{itemize}

\begin{itemize}
\item {\em Domain-incremental learning} where each task shares the same target set, \ie, $\mathcal{Y}_1 = \mathcal{Y}_2 = \cdots = \mathcal{Y}_T$.
\end{itemize}

\subsection{Evaluation Metrics}\label{sec:suppl_metric}
Formally, suppose the model is conducted for $N$ tasks and let $A_{i.j}$ denote the classification accuracy evaluated on the test set of the task $i$ after the incremental learning of the $j$-th task is $A_{i.j}$. Our method is extensively evaluated by three commonly used metrics:

\begin{itemize}
\item {\em Last-step accuracy (Last)} which measures the overall performance at last:
\begin{equation}
\begin{aligned}
    Last&=\frac{1}{N}\sum_{i=1}^{N} A_{i,N} \\
\end{aligned}
\end{equation}
\end{itemize}

\begin{itemize}
\item {\em Average incremental accuracy (Avg)} which measure the performance evolution along the learning trajectory:
\begin{equation}
\begin{aligned}
    Avg&=\frac{1}{N}\sum_{j=1}^{N}(\frac{1}{j}\sum_{i=1}^{j} A_{i,j}) \\
\end{aligned}
\end{equation}
\end{itemize}

\begin{itemize}
\item {\em Forgetting rate (Forget)} which measures the degree of forgetting on learned tasks:
\begin{equation}
\begin{aligned}
    Forget&=\frac{1}{N-1}\sum_{i=1}^{N-1}\max\{A_{i,1},\cdots,A_{i,N-1}\} - A_{i,N} \\
\end{aligned}
\end{equation}
\end{itemize}

Besides, following~\cite{anatomy}, we perform a subspace similarity analysis to measure the representation drifting. Given the input from the same task, let $\mathbf{F}_t, \mathbf{F}_{t'}\in \mathbb{R}^{n\times d}$ denote the output of the encoder after the $t$-th task and after the $t'$-th task ($t'>t$), respectively. We compute the PCA decomposition of $\mathbf{F}_{t}$, \ie, the eigenvectors $(v_1, v_2,\cdots)$ of $\mathbf{F}_{t}^{\top}\mathbf{F}_{t}$. Let $\mathbf{V}_{k,t}$ are the top-$k$ principal directions of $\mathbf{F}_t$, and $\mathbf{V}_{k,t'}$ the corresponding matrix for $\mathbf{F}_{t'}$. The representation drifting from the $t$-th task to the $t'$-th can be defined as:
\begin{equation}
    \mathrm{RepreDrift}_k(\mathbf{F}_t, \mathbf{F}_{t'}) = 1 - \frac{1}{k}\|\mathbf{V}_{k,t}^T \mathbf{V}_{k,t'} \|_{F}^2.
\end{equation}
$\frac{1}{k}\|\mathbf{V}_{k,t}^T \mathbf{V}_{k,t'} \|_{F}^2$ measures the similarity of the subspaces spanned by $\mathbf{F}_t$ and $\mathbf{F}_{t'}$. The smaller the similarity between the subspaces at task $t$ and $t'$, the greater the representation drifting.

\section{Hyperparameter details}\label{sec:suppl_hyper}
We provide the detailed hyperparameters of class-incremental learning, task-incremental learning, and domain-incremental experiments in Sec.~\ref{sec:suppl_cls_hyper}, Sec.~\ref{sec:suppl_task_hyper} and Sec.~\ref{sec:suppl_domain_hyper}, respectively.

\subsection{Class-incremental learning}\label{sec:suppl_cls_hyper}
For CNN-based methods~\cite{aanet,cwd,lucir,bic,il2m}, we employ the SGD optimizer~\cite{sgd} with an initial learning rate of 0.1, a momentum of 0.9, and a batch size of 128. In the experiments performed on CIFAR100, all models are trained for 160 epochs within each task, with the learning rate decreased by a factor of 10 at the 80-th and 120-th epochs. For ImageNet100, all models are trained for 90 epochs within each task, with the learning rate reduced by a factor of 10 at the 30-th and 60-th epochs.

For ViT-based methods such as DyTox~\cite{dytox}, we follow the original hyperparameters. We train the model for 500 epochs per task with Adam~\cite{adam} with a learning rate of 5e-4, including 5 epochs of warmup. At the end of each task (except the first), we finetune the model for 20 epochs with a learning rate of 5e-5 on a balanced dataset.

\subsection{Task-incremental learning}\label{sec:suppl_task_hyper}
The learning rate starts from 1e-4 and decays at epochs 30 and 60 with a multiplier of 0.1. The total epochs are 80. The batch size is set to 32. The regularization coefficient of EWC~\cite{ewc}, MAS~\cite{mas} and SI~\cite{si} are set to 100, 0.1 and 10, respectively.

\subsection{Domain-incremental learning}\label{sec:suppl_domain_hyper}
We use the Adam~\cite{adam} optimizer with an initial learning rate 0.001, and a batch size of 128. The epochs are 80 and the learning rate is decay by 10 at the 40-th and 60-th epochs. The regularization coefficients of EWC~\cite{ewc}, MAS~\cite{mas}, SI~\cite{si} and GEM~\cite{gem} are set to 100, 0.1, 0.3 and 5, respectively.

\begin{table}[t]
  \centering
  \resizebox{0.5\textwidth}{!}{
  {
    \scriptsize
  \begin{tabular}{p{0.3cm}|lccc}
  \toprule
   \# & Template & Avg ($\uparrow$) & Last ($\uparrow$) & $\mathcal{F}$ ($\downarrow$) \\
    \midrule
    \demph{1} & \demph{Baseline} & \demph{65.9\%} & \demph{55.8\%} & \demph{24.9\%} \\
    \midrule
    2 & \texttt{\{object\}} & 67.5\% & 57.1\% & 22.5\% \\
    3 & \texttt{a photo of a \{object\}} & 67.5\% & 57.5\% & 22.5\% \\
    4 & Templates ensemble~\cite{clip} & 67.2\% & 57.9\% & 21.7\% \\
  \bottomrule
\end{tabular}
  }
}
\caption{Comparison with different prompting techniques on CIFAR-100 under class-incremental setting $B=50, C=10$.}
\label{tab:suppl_temp}
\end{table}
\section{Additional Experiments Analysis}\label{sec:suppl_results}
In Sec.~\ref{sec:suppl_class} and Sec.~\ref{sec:suppl_few}, we present additional results on class-incremental learning and few-shot class-incremental learning experiments, respectively. Moreover, in Sec.~\ref{sec:suppl_temp}, we offer an analysis of the prompting technique.

\subsection{Class-incremental Learning}\label{sec:suppl_class}
In the main manuscript, Table 1 presents the results of class-incremental learning (CIL) experiments on CIFAR100 under the setting where the number of base classes ($B$) equals the number of incremental classes ($C$). To provide further insights, we supplement additional results under the setting where $B=50$ in Tab.~\ref{tab:suppl_cifar}. The results show that our proposed method consistently and significantly improves the performance across all metrics under the $B=50$ setting. These results provide further evidence of the effectiveness of our approach in various CIL settings.

\subsection{Few-shot Class-incremental Learning}\label{sec:suppl_few}
The results of few-shot class-incremental learning are displayed in Figure 7 of the main manuscript. To provide more quantitative results, we also present them in Tab.~\ref{tab:suppl_few}. It is evident that our proposed approach consistently achieves significant performance gains, with or without buffers. These findings provide further evidence that our method facilitates effective knowledge transfer from the initial well-learned task.

\subsection{Prompting Technique}\label{sec:suppl_temp}
Prompting~\cite{clip} is a widely used technique to transfer knowledge from pretrained language models. In Tab.~\ref{tab:suppl_temp}, we compare three different settings for prompting. In setting \#1, we used the category name as input without any additional templates. In setting \#2, we used the template \texttt{a photo of a \{object\}}. Finally, in setting \#3~\cite{clip}, we averaged the results of 80 different templates. Our results show that the use of templates can slightly ease the forgetting, which we attribute to the fact that the template ensemble enhances the stability of the generated features and reduces the effect of noise. These findings highlight the robustness and generalizability of our approach.

\begin{figure}[ht]
\scriptsize
\begin{multicols}{5}
\raggedcolumns

\texttt{apple} \\
\texttt{aquarium fish} \\
\texttt{baby} \\
\texttt{bear} \\
\texttt{beaver} \\
\texttt{bed} \\
\texttt{bee} \\
\texttt{beetle} \\
\texttt{bicycle} \\
\texttt{bottle} \\
\texttt{bowl} \\
\texttt{boy} \\
\texttt{bridge} \\
\texttt{bus} \\
\texttt{butterfly} \\
\texttt{camel} \\
\texttt{can} \\
\texttt{castle} \\
\texttt{caterpillar} \\
\texttt{cattle} \\
\texttt{chair} \\
\texttt{chimpanzee} \\
\texttt{clock} \\
\texttt{cloud} \\
\texttt{cockroach} \\
\texttt{couch} \\
\texttt{crab} \\
\texttt{crocodile} \\
\texttt{cup} \\
\texttt{dinosaur} \\
\texttt{dolphin} \\
\texttt{elephant} \\
\texttt{flatfish} \\
\texttt{forest} \\
\texttt{fox} \\
\texttt{girl} \\
\texttt{hamster} \\
\texttt{house} \\
\texttt{kangaroo} \\
\texttt{computer keyboard} \\
\texttt{lamp} \\
\texttt{lawn mower} \\
\texttt{leopard} \\
\texttt{lion} \\
\texttt{lizard} \\
\texttt{lobster} \\
\texttt{man} \\
\texttt{maple tree} \\
\texttt{motorcycle} \\
\texttt{mountain} \\
\texttt{mouse} \\
\texttt{mushroom} \\
\texttt{oak tree} \\
\texttt{orange} \\
\texttt{orchid} \\
\texttt{otter} \\
\texttt{palm tree} \\
\texttt{pear} \\
\texttt{pickup truck} \\
\texttt{pine tree} \\
\texttt{plain} \\
\texttt{plate} \\
\texttt{poppy} \\
\texttt{porcupine} \\
\texttt{possum} \\
\texttt{rabbit} \\
\texttt{raccoon} \\
\texttt{ray} \\
\texttt{road} \\
\texttt{rocket} \\
\texttt{rose} \\
\texttt{sea} \\
\texttt{seal} \\
\texttt{shark} \\
\texttt{shrew} \\
\texttt{skunk} \\
\texttt{skyscraper} \\
\texttt{snail} \\
\texttt{snake} \\
\texttt{spider} \\
\texttt{squirrel} \\
\texttt{streetcar} \\
\texttt{sunflower} \\
\texttt{sweet pepper} \\
\texttt{table} \\
\texttt{tank} \\
\texttt{telephone} \\
\texttt{television} \\
\texttt{tiger} \\
\texttt{tractor} \\
\texttt{train} \\
\texttt{trout} \\
\texttt{tulip} \\
\texttt{turtle} \\
\texttt{wardrobe} \\
\texttt{whale} \\
\texttt{willow tree} \\
\texttt{wolf} \\
\texttt{woman} \\
\texttt{worm} \\
  \end{multicols}
  \caption{The categories of CIFAR100.}
  \label{fig:suppl_cate_cifar}
\end{figure}

\begin{figure}[h]
\scriptsize
\begin{multicols}{4}
\raggedcolumns

\texttt{Alarm Clock} \\
\texttt{Backpack} \\
\texttt{Batteries} \\
\texttt{Bed} \\
\texttt{Bike} \\
\texttt{Bottle} \\
\texttt{Bucket} \\
\texttt{Calculator} \\
\texttt{Calendar} \\
\texttt{Candles} \\
\texttt{Chair} \\
\texttt{Clipboards} \\
\texttt{Computer} \\
\texttt{Couch} \\
\texttt{Curtains} \\
\texttt{Desk Lamp} \\
\texttt{Drill} \\
\texttt{Eraser} \\
\texttt{Exit Sign} \\
\texttt{Fan} \\
\texttt{File Cabinet} \\
\texttt{Flipflops} \\
\texttt{Flowers} \\
\texttt{Folder} \\
\texttt{Fork} \\
\texttt{Glasses} \\
\texttt{Hammer} \\
\texttt{Helmet} \\
\texttt{Kettle} \\
\texttt{Keyboard} \\
\texttt{Knives} \\
\texttt{Lamp Shade} \\
\texttt{Laptop} \\
\texttt{Marker} \\
\texttt{Monitor} \\
\texttt{Mop} \\
\texttt{Mouse} \\
\texttt{Mug} \\
\texttt{Notebook} \\
\texttt{Oven} \\
\texttt{Pan} \\
\texttt{Paper Clip} \\
\texttt{Pen} \\
\texttt{Pencil} \\
\texttt{Postit Notes} \\
\texttt{Printer} \\
\texttt{Push Pin} \\
\texttt{Radio} \\
\texttt{Refrigerator} \\
\texttt{Ruler} \\
\texttt{Scissors} \\
\texttt{Screwdriver} \\
\texttt{Shelf} \\
\texttt{Sink} \\
\texttt{Sneakers} \\
\texttt{Soda} \\
\texttt{Speaker} \\
\texttt{Spoon} \\
\texttt{TV} \\
\texttt{Table} \\
\texttt{Telephone} \\
\texttt{ToothBrush} \\
\texttt{Toys} \\
\texttt{Trash Can} \\
\texttt{Webcam} \\
  \end{multicols}
  \caption{The categories of OfficeHome.}
  \label{fig:suppl_cate_office}
\end{figure}

\newpage
\begin{figure}[t!]
\scriptsize
\begin{multicols}{2}
\raggedcolumns

\texttt{eastern hog-nosed snake} \\
\texttt{rooster} \\
\texttt{wardrobe} \\
\texttt{corkscrew} \\
\texttt{isopod} \\
\texttt{beaver} \\
\texttt{acorn} \\
\texttt{goldfinch} \\
\texttt{Siamese cat} \\
\texttt{chiffonier} \\
\texttt{bittern bird} \\
\texttt{screw} \\
\texttt{Cairn Terrier} \\
\texttt{valley} \\
\texttt{lens cap} \\
\texttt{Brittany dog} \\
\texttt{Appenzeller Sennenhund} \\
\texttt{entertainment center} \\
\texttt{Greater Swiss Mountain Dog} \\
\texttt{Band-Aid} \\
\texttt{dhole} \\
\texttt{sea anemone} \\
\texttt{ice cream} \\
\texttt{threshing machine} \\
\texttt{bell or wind chime} \\
\texttt{sunglasses} \\
\texttt{can opener} \\
\texttt{microphone} \\
\texttt{quail} \\
\texttt{brussels griffon} \\
\texttt{computer keyboard} \\
\texttt{hand-held computer} \\
\texttt{eel} \\
\texttt{Norwegian Elkhound} \\
\texttt{mailbox} \\
\texttt{leopard} \\
\texttt{mitten} \\
\texttt{Cocker Spaniel} \\
\texttt{split-rail fence} \\
\texttt{dowitcher} \\
\texttt{tennis ball} \\
\texttt{Afghan Hound} \\
\texttt{parking meter} \\
\texttt{snow leopard} \\
\texttt{spiny lobster} \\
\texttt{monarch butterfly} \\
\texttt{hook} \\
\texttt{drumstick} \\
\texttt{toilet paper} \\
\texttt{sawmill} \\
\texttt{silver salmon} \\
\texttt{remote control} \\
\texttt{chain mail} \\
\texttt{swim trunks / shorts} \\
\texttt{white stork} \\
\texttt{teddy bear} \\
\texttt{moped} \\
\texttt{horse chestnut seed} \\
\texttt{holster} \\
\texttt{ping-pong ball} \\
\texttt{purse} \\
\texttt{indigo bunting} \\
\texttt{wolf spider} \\
\texttt{lighthouse} \\
\texttt{sturgeon} \\
\texttt{toaster} \\
\texttt{Arctic fox} \\
\texttt{doormat} \\
\texttt{southern black widow} \\
\texttt{high-speed train} \\
\texttt{vending machine} \\
\texttt{cricket insect} \\
\texttt{longhorn beetle} \\
\texttt{African rock python} \\
\texttt{red wine} \\
\texttt{assault rifle} \\
\texttt{carbonara} \\
\texttt{CRT monitor} \\
\texttt{candy store} \\
\texttt{academic gown} \\
\texttt{cannon} \\
\texttt{music speaker} \\
\texttt{African wild dog} \\
\texttt{farm plow} \\
\texttt{koala} \\
\texttt{crutch} \\
\texttt{Groenendael dog} \\
\texttt{Norwich Terrier} \\
\texttt{cardboard box / carton} \\
\texttt{combination lock} \\
\texttt{candle} \\
\texttt{Windsor tie} \\
\texttt{pan flute} \\
\texttt{rose hip} \\
\texttt{small white butterfly} \\
\texttt{space shuttle} \\
\texttt{Chow Chow} \\
\texttt{wool} \\
\texttt{ring binder} \\
\texttt{alligator lizard} \\
  \end{multicols}
  \caption{The categories of ImageNet100.}
  \label{fig:suppl_cate_i100}
\end{figure}

\end{document}